\pdfoutput=1

\documentclass[journal]{IEEEtran}

\usepackage{cite}
\usepackage[final]{graphicx}
\usepackage{graphicx}
\usepackage{bm}
\usepackage[linesnumbered,ruled,vlined]{algorithm2e}
\SetAlFnt{\small}
\SetAlCapFnt{\small}
\SetAlCapNameFnt{\small}

\usepackage{amsmath}
\usepackage{setspace} 
\usepackage{mathtools}
\usepackage{mathrsfs}
\usepackage{amsfonts}
\usepackage{amsmath}
\usepackage{epsfig}
\usepackage{amssymb}
\usepackage{amsmath}
\usepackage{dsfont}
\usepackage{enumitem}  
\usepackage{caption}
\usepackage{subcaption}
\usepackage{verbatim}
\usepackage[usenames,dvipsnames]{color}
\usepackage[colorlinks=true]{hyperref}
\usepackage[hang, flushmargin]{footmisc}         
\usepackage{footnotebackref}
\usepackage{mathtools}
\usepackage{stackengine}
\usepackage{pagecolor}
\usepackage{balance}

\newcommand*{\Scale}[2][4]{\scalebox{#1}{$#2$}}%
\newcommand{\defeq}{\vcentcolon=}


\DeclareMathOperator*{\argmin}{argmin}
\DeclareMathOperator*{\argmax}{argmax}
\newcommand*{\argminl}{\argmin\limits}

\let\oldnl\nl%
\newcommand{\nonl}{\renewcommand{\nl}{\let\nl\oldnl}}%

\title{Imaging with SPADs and DMDs: Seeing through Diffraction-Photons}

\author{{Ibrahim Alsolami and Wolfgang Heidrich,~\IEEEmembership{Senior~Member,~IEEE}}}

\IEEEpubid{\begin{minipage}{\textwidth}\ \\[50pt] \centering
\texttt{\textbf{DOI: 10.1109/TIP.2019.2941315} \textbf{\copyright IEEE}}\\
 I. Alsolami and W. Heidrich, ``Imaging with SPADs and DMDs: Seeing through Diffraction-Photons,"  \textit{IEEE Transactions on Image Processing}, vol. 29, pp. 1440-1449, 2020.
\end{minipage}}

\begin{document}
\maketitle

\begin{abstract}

This paper addresses the problem of imaging in the presence of diffraction-photons. Diffraction-photons arise from the low contrast ratio of DMDs ($\sim$\,$1000{:}1$), and very much degrade the  quality of images captured by SPAD-based  systems.

Herein, a joint illumination-deconvolution scheme is designed to overcome diffraction-photons, enabling the acquisition of intensity and depth images. Additionally, a proof-of-concept experiment is conducted to demonstrate the viability of the designed scheme. It is shown that by co-designing the illumination and deconvolution phases of imaging, one can substantially overcome diffraction-photons.

\end{abstract}


\section{Introduction}
\label{sec:intro}

Today, imaging with single-photon avalanche diodes (SPADs) is rapidly gaining attention. The high sensitivity and temporal resolution of SPADs is finding favor in a variety of applications, ranging from light-in-flight recording~\cite{Gariepy}  to  imaging around corners~\cite{NLOS}.

Traditionally, photon detection has relied on photomultiplier tubes (PMTs). PMTs have a fast response time and provide a high amplification gain, yet they are fairly large vacuum tubes and thus not well-suited for cameras.

Nowadays, SPADs are increasingly becoming the preferred solid-state device for photon detection. Low dark counts ($<25$ Hz) and high quantum efficiencies (up to $\sim50\%$)~\cite{MPD} are among the attractive features of SPADs.

SPAD cameras are, however, still in their infancy. Currently, commercially available SPAD cameras have  low pixel resolutions ($\sim32\times64$ pixels). To increase the spatial resolution of single-pixel SPAD cameras, raster-scanning, whereby pixel values of a scene are sequentially acquired, is typically used.

Raster-scanning is commonly achieved  by means of galvanometric mirrors. In this approach,  light is steered  to $x/y$ positions of a scene  mechanically via a pair of mirrors. Such mechanical systems, however,  are cumbersome.
 
In contrast to single-point illumination in  galvanometric  systems, digital micro-mirror devices (DMDs) can project optical patterns. A DMD is a spatial light modulator (SLM), comprising  a matrix of micro-mirrors integrated with a semiconductor chip. When light shines on a DMD, high-resolution ($912 \times 1140$ pixels)~\cite{TI} optical patterns are projected; in our experiment, we use these optical patterns to illuminate a scene.

\begin{figure}[!t]
\centering
\includegraphics[scale=1.5]{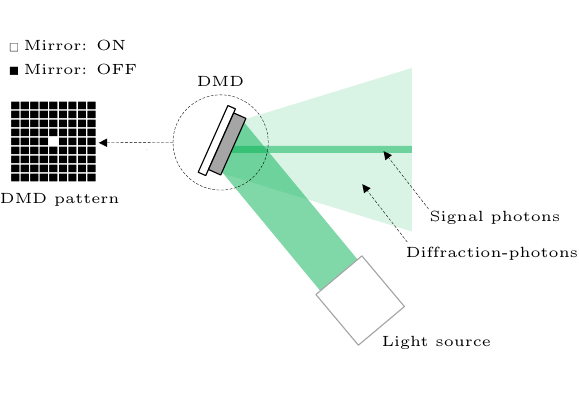}
\caption{Simplified sketch of diffraction-photons emanating from a DMD.}
\label{Fig:Dfi}
\end{figure}

A challenging drawback of DMD-based projection systems, compared with their galvanometric counterparts, is the presence of scattered light. Light diffracted by edges of DMD  structures, such as mirrors, gives rise to scattered light; this light can undergo multiple internal reflections within the DMD and  exit as a wide cone of light, Fig.~\ref{Fig:Dfi}. Due to scattered light, scene locations that would ideally be masked by DMD mirrors in the off-state are illuminated; this imperfection hinders the collection of quality images.

The degree to which a DMD system  rejects light is quantified by the contrast ratio\footnote{The full-on/full-off contrast ratio is defined as  $CR= \frac{L_\text{on}}{L_\text{off}}$, where $L_\text{on}$  is the measured luminance when all DMD mirrors are in the on-state, resulting in a fully illuminated screen, and  $L_\text{off}$ is the  measured luminance when all mirrors are in the off-state.}.  The higher the contrast ratio, the lower the impact of scattered light on image quality. We shall refer to photons originating from a DMD's scattered light as \textit{diffraction-photons}, Fig.~\ref{Fig:Dfi}.

There are further challenges. For example, increasing the illumination power may not always be a feasible option for enhancing image quality, such as when imaging sensitive biological samples~\cite{Handbook}. Moreover, it is desirable to operate an imaging system in a gateless manner in order to reduce the number of measurements.

This study aims to overcome diffraction-photons and provide an alternative to raster-scanning. Using a SPAD, a DMD, and a co-designed illumination-deconvolution scheme, we experimentally demonstrate a system capable of acquiring intensity and depth images in an environment tainted by diffraction-photons.

\bigskip
The contributions of this study are as follows:
\begin{itemize}
\smallskip

\item \textbf{Design}: we design a scheme, tailored to SPAD/DMD-based imaging systems, for capturing intensity and depth images (Sections~\ref{sec:Ill} and \ref{sec:If}).

\item \textbf{Theory}: we formulate image-formation models for our  proposed scheme and  describe techniques  for recovering  images from them (Section~\ref{sec:If}).

\item \textbf{Experiment}: we experimentally demonstrate the viability of our imaging system (Section~\ref{sec:d}).

\end{itemize}

\bigskip

The central idea of this paper can be summarized as follows:

 \smallskip
\textit{Project overlapping illumination blocks to overcome diffraction-photons, and let  deconvolution algorithms untangle pixel values.}

\medskip

The remainder of this paper is organized into seven sections:  In Section \ref{sec:sm}, we  introduce some preliminaries. In Section \ref{sec:rw}, we review prior work. In Section \ref{sec:Ill}, we describe a method for scene illumination, and we then present our image-formation models in Section \ref{sec:If}. In Section \ref{sec:es}, we describe details of our experimental setup. In Section \ref{sec:d}, we present and discuss our experimental results. In Section \ref{sec:c}, we draw conclusions from our findings and outline future research directions.

\section{Preliminaries}
\label{sec:sm}

In this section, we introduce some principles that will serve as starting points for the subsequent sections. This section is based in part  on~\cite{System}.

Each pixel, $i$, of the scene has a depth $z_i$ and reflectivity $\kappa_i$, and is illuminated $N_r$ times by a laser pulse with a waveform $s(t)$. The  rate of photons impinging on a SPAD is given by

\begin{equation} \label{eq:rate}
r_i(t)=\kappa_{i} s(t-2z_{i}/c)+n_a \qquad   \text{photons/sec}
\end{equation}

\noindent where $c$ is the speed of light and $n_a$ is the average photon rate due to ambient light.

The quantum efficiency of the SPAD lowers the photon rate by a factor of $\eta$, and dark counts, $n_d$, of the SPAD are added:

\begin{equation} \label{eq:2}
p_{i}(t)=\eta\kappa_{i} s(t-2z_{i}/c)+ \eta n_a+n_d \qquad  \text{photons/sec}.
\end{equation}

A time-correlated single-photon counting (TCSPC)  system quantizes the observation interval, $T_b$, into $m$ time-bins of duration $\Delta $. Accordingly, the average number of photon counts in a time-bin (per illumination)  is

\begin{equation} \label{eq:4}
 \lambda_{i}^j=\int_{\Delta (j-1)}^{\Delta j } p_{i}(t) dt \qquad  \qquad  \text{photons}
 \end{equation}
where $j$ is the index of a  time-bin, $ j=1,2,\dots,m$.

After $N_r$ illuminations, the number of photons, $y_j$, detected in a time-bin $j$ draws values from a Poisson distribution:

\begin{equation} \label{eq:Poisson_Time-bin}
y_{j}\sim \text{Poisson} \bigg(N_r  \lambda_{i}^j\bigg) \qquad  \qquad  \text{photons}.
\end{equation}

The signal-to-background ratio (SBR), a measure of signal fidelity, is defined as follows: SBR$=\frac{\lambda_s}{\lambda_n}$, where $\lambda_s$ and $\lambda_n$ are the mean number of  signal and noise photons, respectively. Moreover, deadtime is defined as the time a SPAD needs to reset to its initial state after an avalanche event occurs. During this period of time, arriving photons will go undetected.

\section{Prior Work}
\label{sec:rw}

Based on the method of acquisition employed, the literature can be categorized into the following classes:

\medskip
\textbf{Raster-Scanning:}
In raster-scanning pixel values are  sequentially acquired. Typically, a pair of galvanometric mirrors or a motorized translation stage is used to obtain a scene's spatial content~\cite{First_Photon,Stanford, Photonics, Shin}.

Our imaging system and systems in this category require the same number of measurements to reconstruct an intensity and depth image. The main problems, however, associated with mechanical scanning systems such as galvanometric mirrors or motorized translation stages is that they are substantially large, require high power levels, and in the case of galvanometric mirrors, suffer from geometric distortion~\cite{ThorlabsD}. A solution to these problems is to employ a DMD, which we use in our experiment.

\medskip
\textbf{Time-Gated Compressive Sensing:}
In~\cite{Sweep2, Sweep3, Sweep4}, 3D imaging systems based on compressive sensing were demonstrated. Using range gating, time-slices of a scene are obtained, from which depth images are reconstructed.

While such systems can be used to reduce the impact of diffraction-photons, a full time sweep is required to construct a 3D image, as gating is needed to distinguish objects that fall within the same range interval.

What sets our imaging system apart from others in this category is that it is gateless, hence requiring fewer measurements to reconstruct a 3D image. The number of  measurements in our imaging system is equal to the number of pixels, $n$, as opposed to  $\left\lfloor\dfrac{n}{d}\right\rfloor\times m$ for gated compressed sensing, where $d$ is  a selected constant for a given image sparsity/compressibility (typically in the range between 2 and 4) and $m$ is the number of time-bins ($m\gg d$).

\medskip
\textbf{Gateless Compressive Sensing:}
Methods of circumventing the need for time gating have been proposed. In~\cite{Andrea}, an imaging system able to construct a $64 \times 64$ depth image in a gateless manner  was demonstrated.  Prior to recovering a scene's spatial information, parametric deconvolution is first used to estimate the set of depths present in scene. Once the set of depths is determined, a compressive sensing approach is used to recover a scene's spatial content.

A major obstacle to adopting the aforementioned  system is the presence of \textit{deadtime-interference}. To give an example, consider the following: an object in the background of a scene may not be visible as photons reflected from it may go undetected due to the deadtime initiated by preceding photons reflected from foreground objects. A way to resolve this obstacle is to use a photo diode operating in the linear-mode (analogue mode)~\cite{Sun}, as opposed to Geiger-mode (photon-counting mode); this, however, will entail a loss in photon detection sensitivity.

In our experiment, we employ a SPAD (Geiger-mode), enabling single-photon detection sensitivity. We also use block illumination (Section~\ref{sec:Ill}), which ameliorates deadtime-interference, allowing our imaging system to operate in gateless manner.

\medskip
\textbf{Epipolar Imaging:}
A relatively new method of image acquisition is epipolar imaging~\cite{Epi1, Epi2, Epi3}. In epipolar imaging, an illumination sheet sweeps through a scene, and an array of pixels on an epipolar plane of illumination detects incoming light. This method of imaging  can potentially overcome diffraction-photons as it limits indirect light. To realize such system, however, some hardware, such as controllable mirrors, is needed to select an epipolar plane. Our imaging system is free of such hardware, making it simpler to operate.


\section{Illumination}
\label{sec:Ill}

\begin{figure}[t!]
\hspace*{.4cm}
\begin{subfigure}{0.15\textwidth}
\centering
\hspace*{-.45cm}\includegraphics[width=1\textwidth]{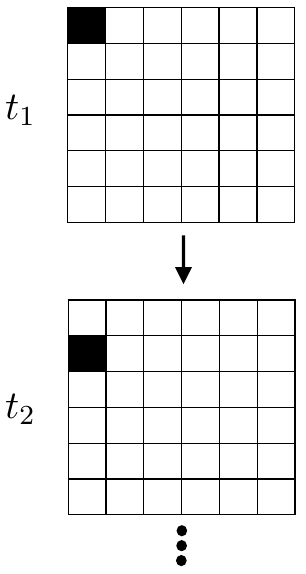}
\caption{}
\label{fig:Rast}
\end{subfigure}%
\hspace{.2cm}
\begin{subfigure}{0.15\textwidth}
\centering
\includegraphics[width=0.78\textwidth]{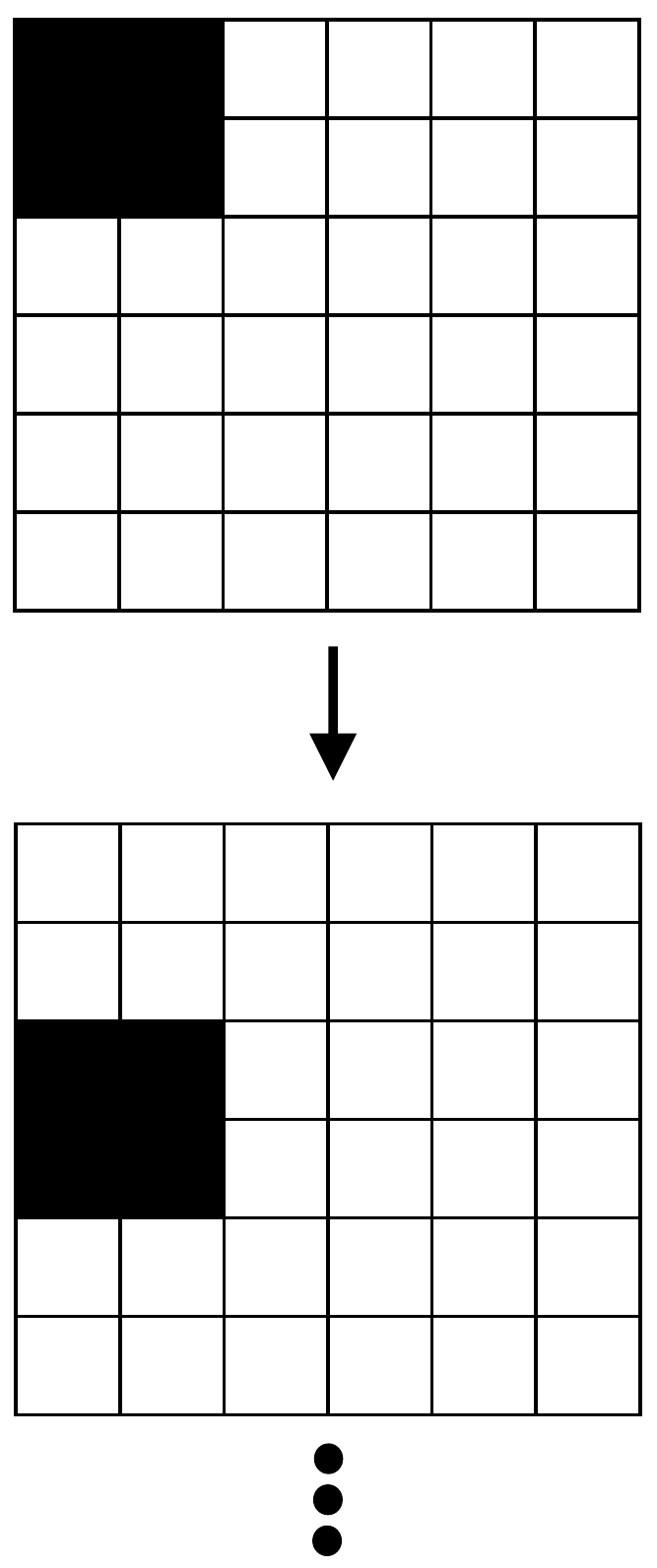}
\caption{}
\label{fig:Lrast}
\end{subfigure}%
\begin{subfigure}{0.15\textwidth}
\centering
\includegraphics[width=0.78\textwidth]{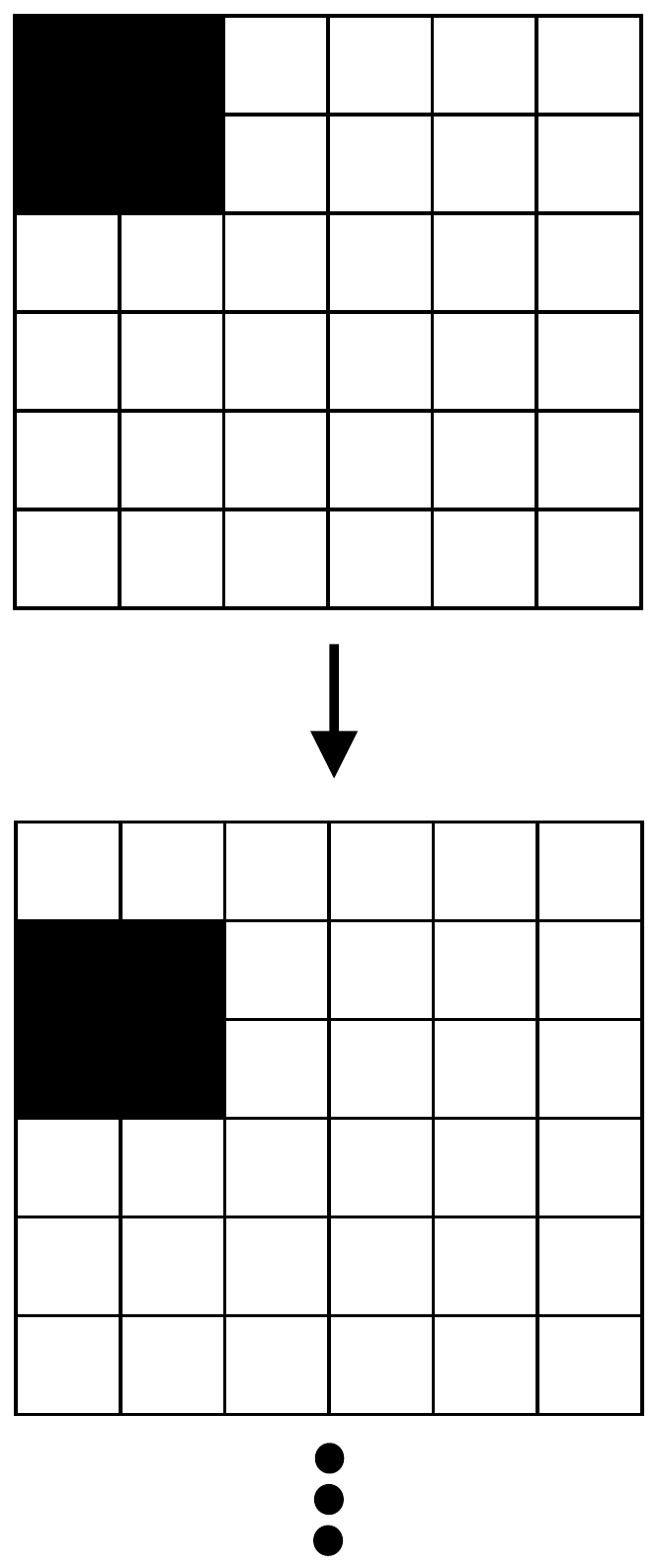}
\caption{}
\label{fig:Ps}
\end{subfigure}%
   \caption{(a) Raster-scanning in a pixel-by-pixel manner. (b) Raster-scanning large patches of the scene. (c) Overlap scanning. In this illustration, an illumination window of size $2 \times 2$ scans the scene in one-pixel steps.}
 \label{Fig:Illumination}
\end{figure}

Trading-off spatial resolution in return for a higher signal power can be  made. At low illumination levels, raster-scanning (Fig.~\ref{fig:Rast}) can yield inaccurate estimations of  intensity and  depth images due to the presence  of diffraction-photons and  low  SBR. To improve  signal power, one can scan large patches of the scene; this, however, comes at the expense of a loss in spatial resolution (Fig.~\ref{fig:Lrast}).

A simple yet effective approach to boosting signal power while maintaining a given spatial resolution is to use an illumination window (for example, of size $w\times w$) that scans the scene in steps of one pixel (Fig.~\ref{fig:Ps}  provides an illustration). This approach improves signal power as more photons are collected and simultaneously retains the native resolution.  Additionally,  deadtime-interference is  alleviated as the optical power is concentrated within an illumination block, where depth values of a scene are notably correlated.

A by-product, however, of using overlapping scanning blocks (Fig.~\ref{fig:Ps}) is a blur artifact as photons from adjacent pixels are collected. In  Section \ref{sec:If}, we overcome this challenge via a deconvolution technique designed  for the problem at hand.

A key advantage of the proposed scheme is that it provides (via optimization) a means to relax the trade-off between spatial resolution and signal power.

Let us now discuss the two prime factors that need to be taken into consideration when selecting an illumination window size, $w \times w$: 

\begin{enumerate}[label=(\roman*)]

\item  \textbf{Deadtime and SPAD saturation:} Increasing the window size results in more signal photons arriving at the SPAD. This, however, can improve performance only to a limited extent, because the deadtime causes the SPAD to saturate at high photon rates.

\item \textbf{Contrast ratio and diffraction-photons:} A DMD with a high contrast ratio requires a relatively small window size, because fewer diffraction-photons are emitted from it; the converse is equally true.

\end{enumerate}

\section{Image Formation}
\label{sec:If}

Using the illumination scheme (Fig.~\ref{fig:Ps}) presented in Section~\ref{sec:Ill}, we now describe approaches to recovering intensity and depth images from photon-arrival events. We first formulate the image reconstruction process as a discrete inverse problem and then solve it by a convex optimization algorithm.

\subsection{Intensity Image}

The goal here is to reconstruct an intensity image,  $\bm {\vec \alpha}_{\text{\tiny opt}}$, with the aid of the proposed illumination scheme (see Fig.~\ref{fig:Ps}). The input--output relationship, in the absence of noise, can be described as follows:
\begin{equation} \label{eq:Iconv}
 h(x,y)\circledast   \alpha(x,y) =v(x,y)
\end{equation}

\noindent where $\circledast $ denotes spatial convolution, $ h(x,y)$ is the point-spread function that describes the proposed illumination scheme (Fig.~\ref{fig:Ps}),  $\alpha(x,y)$ is the latent intensity image,  and $ v(x,y)$ is the observed number of photon counts at pixel $(x,y)$.

Let $\bm {\vec \alpha}=(\alpha_1,\dots,\alpha_n)^T$ denote the vectorized  representations of  image $\alpha(x,y)$---column-wise stacked. Likewise, let $\bm {\vec v}=(v_1,\dots,v_n)^T$ denote the observation vector, where $v_i$ is the total number of photon counts  at the $i^\text{th}$pixel.

The two-dimensional convolution in Eq. \ref{eq:Iconv}  can be expressed as matrix--vector multiplication:

\begin{equation} \label{eq:Multi}
\bm H \bm {\vec \alpha}=\bm {\vec v}
\end{equation}

\begin{equation} \label{eq:Mn}
  \begin{pmatrix} h_{1} &h_2& \dots&h_{n} \\ h_n& h_1&\dots &h_{n-1}\\ \vdots&\vdots&\ddots&\vdots \\h_2 &h_3&\dots&h_{1}  \end{pmatrix}_{\!\!\!\!n \times n} 
   \begin{pmatrix}  \alpha_1   \\ \vdots \\ \alpha_n \end{pmatrix}
= \begin{pmatrix} v_1 \\ \vdots \\ v_n \end{pmatrix}
\end{equation}

\noindent where $h_{i}$ is the DMD reflection coefficient. When a pixel of the scene is within an illumination block (see Fig.~\ref{fig:Ps}), $h_{i}$ takes on a value of one---full reflection. Other pixels of the scene are, however, moderately illuminated due to the low contrast ratio of the DMD: when a DMD mirror is in the off-state, a fraction of the optical power illuminates  the pixels of the scene; we denote this fraction by $\epsilon$. More formally:

\[  \Scale[.9]{h_{i}= \left\{
\begin{array}{ll}
      1,&\text{if~} 0< i~(\text{MOD}{~\Theta})\le w ~~\land~ ~\dfrac{i-i~(\text{MOD}{~\Theta})}{\Theta}+1\le w
      \\
      \epsilon,& \text{otherwise}
\end{array} 
\right.} \]

\medskip

\noindent where $\Theta$ denotes the number of rows of the image, and $w $ is the length of an illumination window of size $w \times w$ (Fig.~\ref{fig:Ps} provides an illustration).  Additionally, let ``$ a~(\text{MOD}{~b })$"  denote  the smallest positive integer ``$q$'' such that $ q \equiv a\pmod b$.

  \bigskip

   The following four steps are performed to recover $\bm {\vec \alpha}$ (Eq.~\ref{eq:Multi}):

  \medskip
  
     \begin{enumerate}

  	 \item \textbf{Variance-stabilizing transformation:}

  	 In Eq.~\ref{eq:Multi}, $\bm{\vec v}$  is a Poisson distributed random vector; to use an $\ell_2$-norm apt for Gaussian noise, we apply an Anscombe transform  to convert  the signal-dependent Poisson noise  of $\bm{\vec v}$ to (approximately) additive Gaussian noise with a constant variance~\cite{Anscombe}:

  	  	  	\begin{equation} \label{eq:Anscombe} 	
							 f(v_i)=2\sqrt{v_i+\frac{3}{8}}~.
						\end{equation}

   The Anscombe transformation, however, breaks the commonly used linear model expressed by Eq.~\ref{eq:Multi}---as $f(v_i)$ is nonlinear operator. We can  circumvent this barrier by first denoising a blurred image (Eq.~\ref{eq:De}), then applying an inverse transform (Eq.~\ref{eq:Inverse}), and finally recovering the latent image (Eq.~\ref{eq:Mnop}).

  \medskip
    	 	     \item \textbf{Denoising:}
    	 	     
The aim here is to   denoise a  blurred image prior to deconvolving it:

\begin{equation}\label{eq:De}
\begin{aligned}
 \vec {\bm b}_{\text{\tiny opt}}&=\argmin_{\vec {\bm b}} \quad \frac{1}{2} \lVert  \vec {\bm b}- f(\bm{\vec v})\lVert^2_2\quad+\quad \mu \Vert \bm D \vec {\bm b}\Vert_1\\
&\text{s.t.} \quad  b_i \geq  2\sqrt{\frac{3}{8}} \quad \quad   \forall  i,~~i= 1,\dots, n
\end{aligned}
\end{equation}
    	 	    
    	 	   \noindent where $\vec {\bm b} \defeq f(\bm H\bm {\vec \alpha})$ and  $\vec {\bm b}_{\text{\tiny opt}}$ is a denoised yet blurred image.
					A constraint  is added because the minimum  value of  Eq.~\ref{eq:Anscombe} is $2\sqrt{\frac{3}{8}} $, which occurs when $f(0)$.

  \bigskip
    \item \textbf{Inverse transformation:}

 Algebraic inversion of  Eq.~\ref{eq:Anscombe} produces a bias; we therefore use a maximum likelihood (ML) transformation~\cite{Foi}:

     \begin{equation} \label{eq:Inverse} \mathcal{I}_\text{ML}: \vec {\bm b}_{\text{\tiny opt}}  \longmapsto   \vec {\bm b}_{\tiny *} \end{equation}

\noindent where $\vec {\bm b}_{\tiny *}$ can be regarded as the denoised version of $\bm{\vec v}$ (Eq.~\ref{eq:Multi}). In this step, a lookup table is used to map the values of $\vec {\bm b}_{\text{\tiny opt}}$ to $\vec {\bm b}_{\tiny *} $. 

\bigskip
 \item \textbf{Deconvolution:}

  We now wish to solve the following system of equations $\bm H\bm {\vec \alpha}= \vec {\bm b}_{\tiny *}$.
Matrix $\bm H$ is ill-conditioned. Accordingly, we consider a regularized least-squares approach to recover our latent image:

\begin{equation}\label{eq:Mnop}
\begin{aligned}
 \bm {\vec \alpha}_{\text{\tiny opt}}&=\argmin_{\bm {\vec \alpha}} \quad \frac{1}{2} \lVert  \bm H\bm {\vec \alpha}-  \vec {\bm b}_{\tiny *}\lVert^2_2\quad+\quad \lambda \Vert \bm D \bm{\vec \alpha}\Vert _1    \\
&\text{s.t.} \quad  \alpha_i \geq  0 \quad \quad   \forall  i,~~i= 1,\dots, n.
\end{aligned}
\end{equation}

 \end{enumerate} 
\vspace{.1cm}

We solve \ref{eq:De} and \ref{eq:Mnop} using the alternating direction method of multipliers (ADMM) algorithm. Details of the ADMM algorithm are provided in the Appendix.

\medskip

\noindent \textbf{Limitation:} A limitation of the procedure described (in steps 1--4) is that a low or high value of $\mu$ in Eq.~\ref{eq:De} may diminish the quality of the image recovered in step 4 (Eq.~\ref{eq:Mnop}), and thus one must carefully select $\mu$. In our experimental results (Section~\ref{sec:d}),  the value of $\mu$ is chosen by inspection: the value of $\mu$ that produces the best result is selected.

\subsection{Depth Image}
The aim here is to obtain a depth image, $\bm{\vec z}$, from photon-arrival events. In the absence of noise, the input--output relationship can be expressed as follows:

\begin{equation} \label{eq:2conv}
h(x,y)\circledast_1 \delta(x,y,t)\circledast_2 s(t) =r(x,y,t).
\end{equation}

\noindent Here, two convolution processes take place:

\begin{enumerate}[label=(\roman*)]
\item $\circledast_1$ is a 2D spatial convolution over $(x,y)$, describing the convolution of the scene with  the proposed overlapping illumination blocks, $h(x,y)$ (Fig.~\ref{fig:Ps} provides an example).
\item $\circledast_2$ is a 1D temporal convolution over $t$, describing the convolution of the scene with the waveform of the illumination pulse, $s(t)$.
\end{enumerate}

\begin{figure} [!t] 
\centering
\begin{subfigure}[b]{\linewidth} 
	\includegraphics[scale=.7]{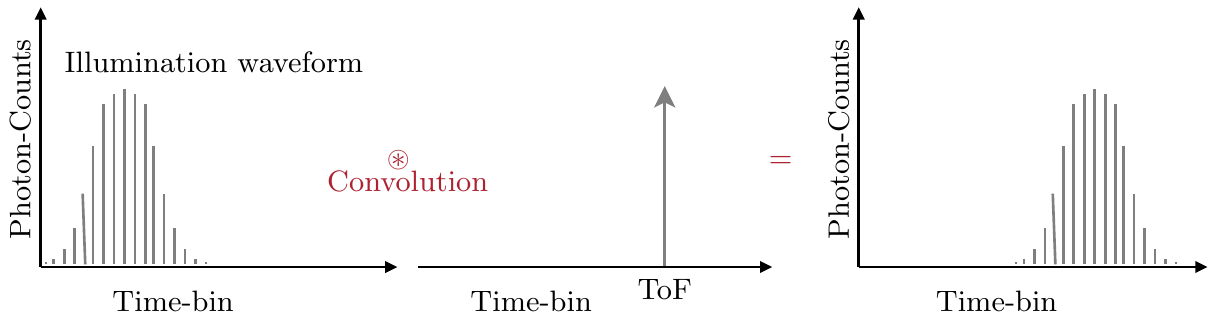}
	
   \caption{}
   \label{fig:IC1} 
\end{subfigure}
\begin{subfigure}[b]{\linewidth} 
\centering
   \includegraphics[scale=1.5]{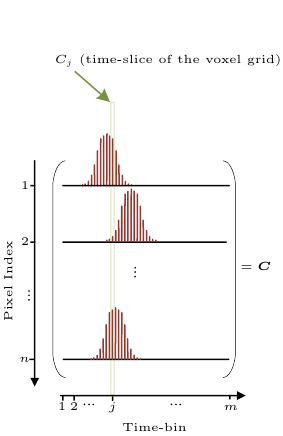}       
   \caption{}
   \label{fig:IC2}
\end{subfigure}

\caption{(a) Convolution of illumination waveform with a depth point of a scene. (b) An illustrative example of the formation of $\bm C$ and $C_j$. Matrix $\bm C$ is the result of convolving the depth at each pixel with the illumination waveform. Each row of $\bm C$ is for a given pixel, and vector $C_j$ is the $j^{\text{th}}$ column of  $\bm C$.} \label{fig:ICa} 
\end{figure}

Each pixel, $i$, of the scene has a depth, $z_i$, value and a corresponding time-of-flight (ToF), $ \frac{2}{c}z_{i}$,  which can be represented by the following impulse response:

\begin{equation} \label{eq:impulse}
\delta\left(t-\frac{2}{c}z_{i}\right), \quad t\in [ 0,T_b)
\end{equation}

 \noindent where $\delta(\cdot)$ is a Dirac function (Fig. \ref{fig:IC1}) and $c$ is the speed of light.  Likewise, in the continuous spatial and temporal domain, function $\delta(x,y,t)$ in Eq.~\ref{eq:2conv} is zero except at the ToF. Additionally, let $r(x,y,t)$ denote the number of photons at a given space-time point $(x,y,t)$.

\vspace{5mm}
\noindent \textbf{Discretization:}
The continuous volume containing the scene is discretized into a voxel grid: the DMD discretizes the scene spatially into $n$ pixels, and the TCSPC module discretizes the scene temporally into $m$ time-bins.  In this voxel grid, a discrete depth image is represented by a vector $\bm{z} \in \{1,\dots,m\}^{n}$.

To keep the notation light, the discrete version of  $\delta(x,y,t)$ in this voxel domain is denoted by $\bm\delta(\bm z)$---akin to Eq.~\ref{eq:impulse}, but now for a complete image---and is given by

\begingroup
\renewcommand*{\arraystretch}{1.3}

\begin{equation} \label{eq:DeltaMat}
  \bm\delta(\bm z)=\begin{pmatrix} 
  \delta_0 \left(t_1,\frac{2}{c}z_1\right)& \dots&\delta_0 \left(t_m,\frac{2}{c}z_1\right) \\ \\[-9pt]
  \delta_0 \left(t_1,\frac{2}{c}z_2\right)& \dots&\delta_0 \left(t_m,\frac{2}{c}z_2\right)\\
 \vdots&\ddots&\vdots\\
 \delta_0 \left(t_1,\frac{2}{c}z_n\right)& \dots&\delta_0 \left(t_m,\frac{2}{c}z_n\right)  \end{pmatrix}_{\!\!n \times m}
\end{equation}
\endgroup

\noindent where $t_k \in \{1,\dots,m\}$ is the $k^{\text{th}}$ time-bin, $c$ is the speed of light,  $z_i$ is the depth at pixel $i$, $m$ is the number of time-bins, $n$ is the total number of pixels, and  $\delta_0(\cdot)$ is a Kronecker delta function defined as

\[ {\delta_0(a,b)= \left\{
\begin{array}{ll}
      0,&\text{if~} a \ne b   \\
       1,&\text{if~} a = b  
\end{array} 
\right.} \]

Each column of matrix  $\bm\delta(\bm z)$ in Eq.~\ref{eq:DeltaMat} represents a time-slice of the voxel grid at a particular time-bin, $t_k$, and each row is allotted to a single pixel of the scene. Additionally, matrix $\bm\delta(\bm z)$  is a binary matrix, and the summation of entries along each row of  $\bm\delta(\bm z)$ has a value of one---as each pixel  can only have a single depth.

The convolution operators in Eq.~\ref{eq:2conv} can be represented  as a matrix  multiplication:
\begin{equation} \label{eq:c4}
\bm H\cdot\bm\delta(\bm z)\cdot \bm S = \bm R\quad
\end{equation}

\noindent where

\begin{equation} \label{eq:multiM4}
 \bm S=\begin{pmatrix} s_{1} &s_2& \dots&s_{m} \\ s_m& s_1&\dots&s_{m-1}\\ \vdots&\vdots&\ddots&\vdots \\s_{2} &s_{3}&\dots&s_{1}  \end{pmatrix}_{\!\!m\times m}.
\end{equation}

\noindent Here, each row of matrix $\bm S$ is a time-histogram of the illumination  waveform and is a circular shift of its preceding row. When a row of $\delta(\bm z)$ is multiplied by  matrix $\bm S$, the illumination waveform is shifted to the ToF (Fig.~\ref{fig:IC1} provides a detailed example).

  In Eq. \ref{eq:c4}, matrix  $\bm R$  is the observed time-histogram at the detector (Fig.~\ref{fig:IRc} provides an illustration). Each row of $\bm R$ is a time-histogram, resulting from convolving the scene with both an illumination window and pulse waveform:
\begin{equation} \label{eq:multiM5}
\bm R= \begin{pmatrix} r_{1,1}  &\cdots& r_{1,m} \\\vdots &\ddots& \vdots \\ r_{n,1}& \cdots& r_{n,m} \end{pmatrix}_{\!\! n \times m}.
\end{equation}

\noindent \textbf{Optimization:}
The inverse problem of recovering $\bm z$ (Eq.~\ref{eq:c4}) can be formulated using the following optimization problem:

 \begin{equation} \label{eq:Dopt}
\bm  z {\text{\tiny opt}}=\argmin_{\bm z} \quad \frac{1}{2} \lVert  \bm H \cdot \bm\delta(\bm z) \cdot \bm S - \bm R\lVert^2_F\quad+\quad \beta \Gamma(\bm z)
\end{equation}

 \noindent where  $ \Gamma(\bm z)$ is a regularizer weighted by $\beta$.  Eq.~\ref{eq:Dopt} is a combinatorial optimization problem, and it is challenging to solve within a reasonable period of time (the search space is $m^n$).
 
A tractable method for inferring depth from Eq.~\ref{eq:Dopt}  can be attained using the following three steps:

 \begin{enumerate}
  
   \vspace{4mm}
 \item  \textbf{Spatial deconvolution:} 
  \vspace{2mm}
 
 In Eq.~\ref{eq:Dopt}, let $\bm C=\bm\delta(\bm z) \cdot \bm S$, and the $j^{\text{th}}$ column of $\bm C$ and $\bm R$ be denoted as $C_j$ and $R_j$, respectively (Figs.~\ref{fig:ICa} and ~\ref{fig:IRc} provide illustrations). Both $C_j$ and $R_j$, are $n$-dimensional vectors: $C_j=(c_1,\dots,c_n)^T$ and $R_j=(r_1,\dots,r_n)^T$, where $j=1,\dots,m$.

The optimization problem for each time-slice (Fig. \ref{fig:IC2}) of the voxel grid is

\begin{equation}\label{eq:CT}
\begin{aligned}
\bar C_j &=\argmin_{ C} \quad \frac{1}{2} \lVert  \bm H C_j- R_j\lVert^2_2\quad+\quad \mu \Vert   \nabla C_j\Vert_{1}\\
&\text{s.t.} \quad  c_i \geq  0 \quad \quad   \forall  i,~~i= 1,\dots, n.
\end{aligned}
\end{equation}

Solving this optimization problem for each $C_j$ yields matrix $\bar {\bm C}=(\bar C_1,\dots,\bar C_m)$. We solve this convex optimization problem using an ADMM algorithm. Details of the ADMM algorithm are provided in  the Appendix.

\begin{figure}[!t] 
    \centering
        \includegraphics[scale=1.55]{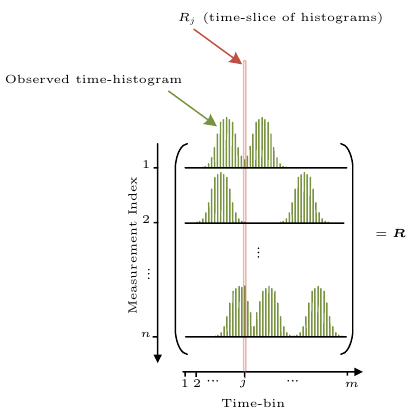}     
    \caption{An illustrative example of the formation of $\bm R$ and $R_j$. Matrix $\bm R$ is the observation matrix, and is the result of convolving the scene with: 1) the proposed overlapping illumination blocks 2) the illumination waveform. 
Each row of $\bm R$ is the observed time-histogram at a given measurement. Vector $R_j$ is the $j^{\text{th}}$ column of $\bm R$, and represents a time-slice of the observed histograms at time-bin $j$.}
    \label{fig:IRc}
\end{figure}

\medskip
 \item  \textbf{Intermediate filtering:}

Prior to temporally deconvolving our image, we apply an  $N^{\text{th}}$-order median filter. 
Experimentation has shown that this filter significantly improves our results.

Let $\bar C_i=(\bar c_1,\dots,\bar c_m)$ denote the $i^{\text{th}}$ row of matrix $\bar {\bm C}$. The $N^{\text{th}}$-order median filter of $\bar c_j$ is

 \begin{equation} \label{eq:mf}
 \hat{c}_j= \text{median}\left(\bar c_{j-\lfloor \frac{N}{2}\rfloor},\dots,\bar c_{j+\lfloor \frac{N}{2}\rfloor}\right), ~\text{for}~j= 1,\dots, m 
 \end{equation}

from which we obtain $\hat{C}_i=(\hat{c}_1,\dots,\hat{c}_m)$.

  \vspace{5mm}
 
 \item  \textbf{Temporal deconvolution:}
  \vspace{2mm}

Using $\hat{C}_i$ from the previous step, we can now determine $\bm{\vec z}$ directly in a per-pixel manner. Vector $\hat{C}_i$ contains the received signal for pixel $i$, which can be regarded as a delayed and scaled  version of the transmitted pulse. The ToF (or signal delay) for each pixel is independently obtained  by cross-correlating $\hat{C}_i$ with the waveform of the transmitted pulse, $s[\, \cdot \,]$, and finding the maximum, as follows:

 \begin{equation} \label{eq:xcross}
 t^{*}_i=\Delta \argmax_{\{0\leq b \leq m-1\}} \sum_{a=1}^{m}  \hat{C}_i[a] s[a-b].
 \end{equation}

For simplicity of notation, $\hat{C}_i[a]$ denotes the $a^\text{th}$ element of vector $\hat{C}_i$. Using Eq.~\ref{eq:xcross}, the depth at pixel, $i$, is  $z_i=\dfrac{c}{2}  t^{*}_i$, from which we obtain $\bm{\vec z}=(z_1,\cdots,z_n)^T$.

 \end{enumerate}

\section{Hardware and Experimental Setup}
\label{sec:es}

\begin{figure}[!t]
\centering
\includegraphics[scale=.5]{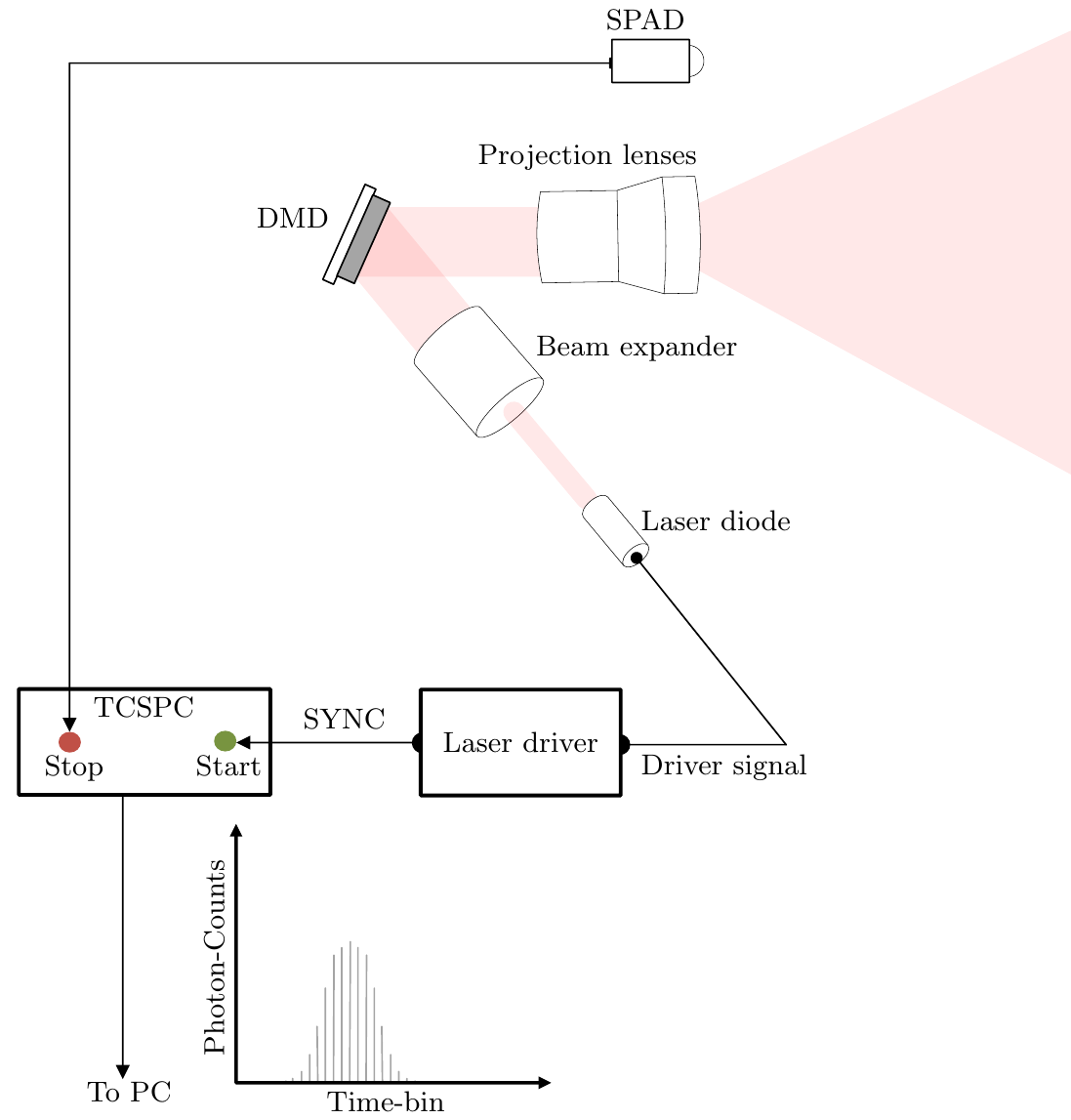}
\caption{Simplified schematic of experimental setup.}
\label{Fig:SETe}
\end{figure}

Fig.~\ref{Fig:SETe} shows a schematic of the experimental setup. The illumination source is a laser diode (LDH-P-C-650, PicoQuant) with a central wavelength of $\sim656$ nm. The laser is controlled by a laser driver  (PDL 828 Sepia II, PicoQuant) and emits pulses with a duration of $\sim 80$ ps (FWHM) and a repetition rate of $70$ MHz.

An optical beam expander enlarges the laser beam by a factor of $5$. The DMD (in a DLP 4500 projector, Texas Instruments) spatially modulates the incoming light and projects it towards the scene. The contrast ratio of the DMD is $\sim$\,$1000{:}1$.

Photons reflected by surfaces of the scene are detected by a SPAD (PDM $20~\mu$m Series, Micro Photon Devices) with a $\sim24$ ps (FWHM) temporal resolution. The SPAD has a $\sim 35\%$ quantum efficiency, $\sim77.8$ ns deadtime, active area diameter of  $\sim20~\mu$m, and dark counts rate of $\sim3.6$ Hz. Photons detected by the SPAD are time-stamped, relative to a sent laser pulse, with a  resolution of $\sim4$ ps by a TCSPC module (PicoHarp 300, PicoQuant).

The scene is spatially sampled by the DMD into $14440$ pixels, forming an image of size $95$ rows by $152$ columns.  The observation interval is divided by the TCSPC module into $1410$ time-bins, each with a duration of $\sim 4$ ps. A total of $5\times10^6$ laser pulses are transmitted for each measurement. The integration time for each measurement is $28.2$ ms.

The scene consists of a ball ($\sim$\,$11$ cm radius) placed in front of a screen. The ball is  $\sim50$ cm away from the SPAD, and the horizontal distance between the SPAD and the background screen is $\sim 72$ cm.
Table~\ref{tab:EXP} provides a summary of the experimental parameters.

\setlength\tabcolsep{4.3pt} %

\begin{table}[!thb]
\footnotesize
\centering
\caption{Experimental parameters}\label{tab:EXP}
\begin{tabular}{l r}
\hline
\noalign{\vskip .7mm}  

Parameter         & Value \\
\noalign{\vskip .7mm}  
\hline\hline

\noalign{\vskip 1.5mm}  
Average dark-count rate of SPAD& $\sim 3.6$~Hz   \\
Center wavelength of laser & $\sim 656~$nm \\ 
Contrast ratio of DMD & $\sim$\,$1000{:}1$\\
Deadtime$^\dag$of SPAD& $\sim77.8$~ns   \\
Diameter of active area of SPAD & $\sim20~\mu$m\\
Integration time per measurement & $28.2$ ms\\
Laser  pulse  duration$^\ddagger$  & $ \sim80$ ps  \\
Number of transmitted laser pulses  per measurement & $5\times10^6$ \\
Number of time-bins & $1410$ \\
Quantum efficiency of SPAD & $\sim35\%$ \\
Repetition rate of  laser pulses & $70$ MHz\\
Timing resolution of SPAD &  $\sim24$ ps\\
Timing resolution of TCSPC module & $\sim4$ ps\\
\noalign{\vskip 1.5mm}

\hline
\noalign{\vskip 1mm}  
\footnotesize $^\dag$Non-extensible deadtime~\cite{DeadTime}. $^\ddagger$Full width at half maximum.
 \end{tabular}
\end{table}


\section{Results and Discussion}
\label{sec:d}

In this section, we present and discuss the results of a proof-of-concept experiment. 
The goal is to demonstrate the ability of the designed illumination-deconvolution scheme (Sections~\ref{sec:Ill} and \ref{sec:If}) to overcome diffraction-photons.

\begin{figure*}[!th]
	\centering
		\includegraphics[scale=.55]{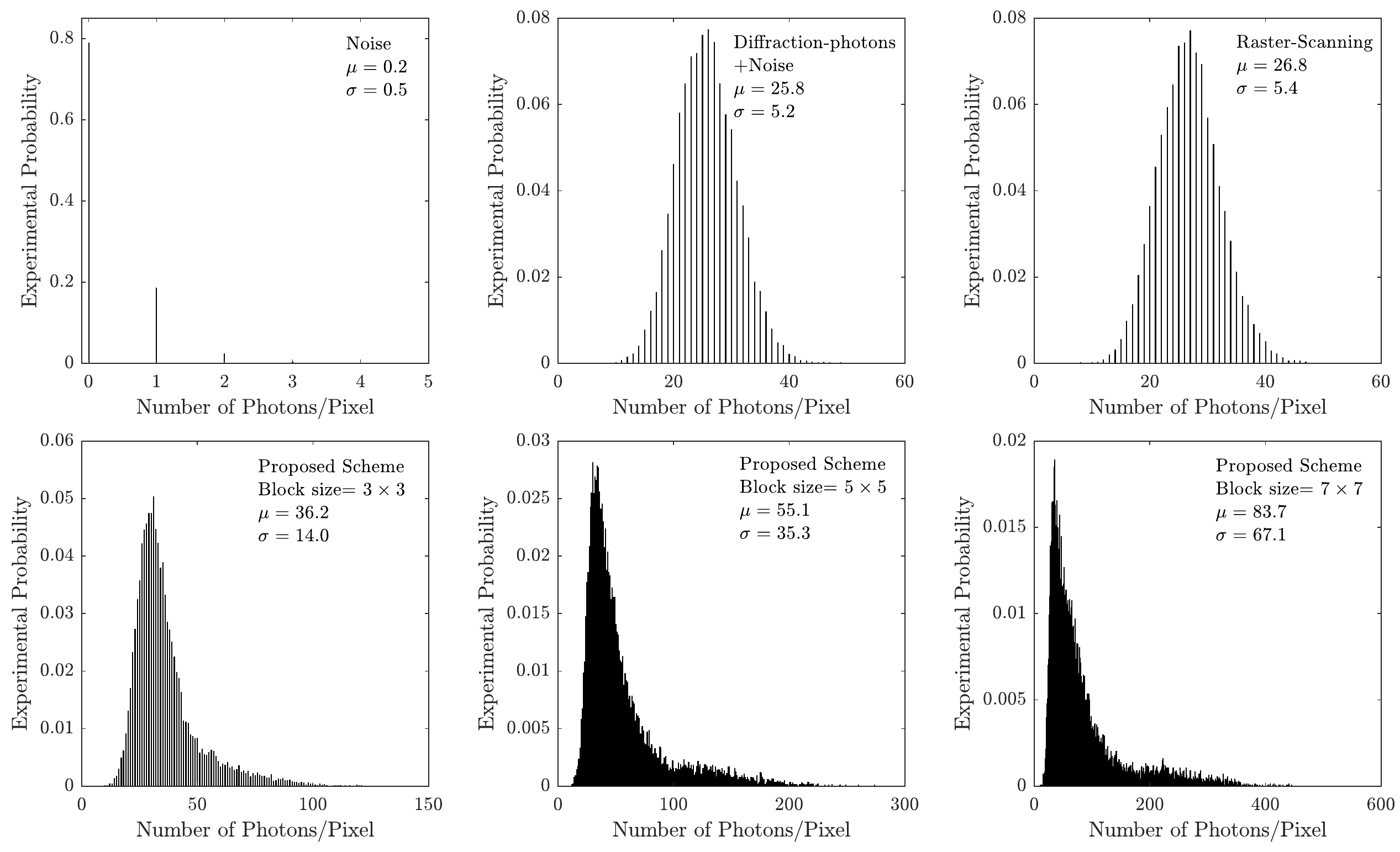}

	\caption{Histograms of experimental data. \textit{Supplementary details:} for raster-scanning and the proposed scheme, the  measured photons/pixel comprise signal, noise  and diffraction photons.}
	\label{Fig:Histograms_All}
\end{figure*}

Fig.~\ref{Fig:Histograms_All} displays histograms of  our experimental data, and the statistics of our measurements are as follows: the average number of noise (ambient + dark) photons was measured and found to be $0.2 \pm 0.5$ photons/pixel.  The average number diffraction-photons with noise  was measured and found to be $ 25.8 \pm 5.2$ photons/pixel. For consistency, we have taken $14440$ samples for all measurements presented in this section---this value is equal to the number of pixels of the sought image.

The average number of photons (signal + noise + diffraction) per pixel for raster-scanning, and illumination blocks of size $3\times 3$, $5\times 5$, and $7\times 7$, was measured and found to be  $26.8 \pm 5.4$, $36.2 \pm 14.0$, $55.1 \pm 35.3$, and $83.7 \pm 67.1$ photons/pixel, respectively.

\begin{figure*} [!htb]
\centering
 \begin{subfigure}[b]{\linewidth} 
\centering
   \includegraphics[scale=.05]{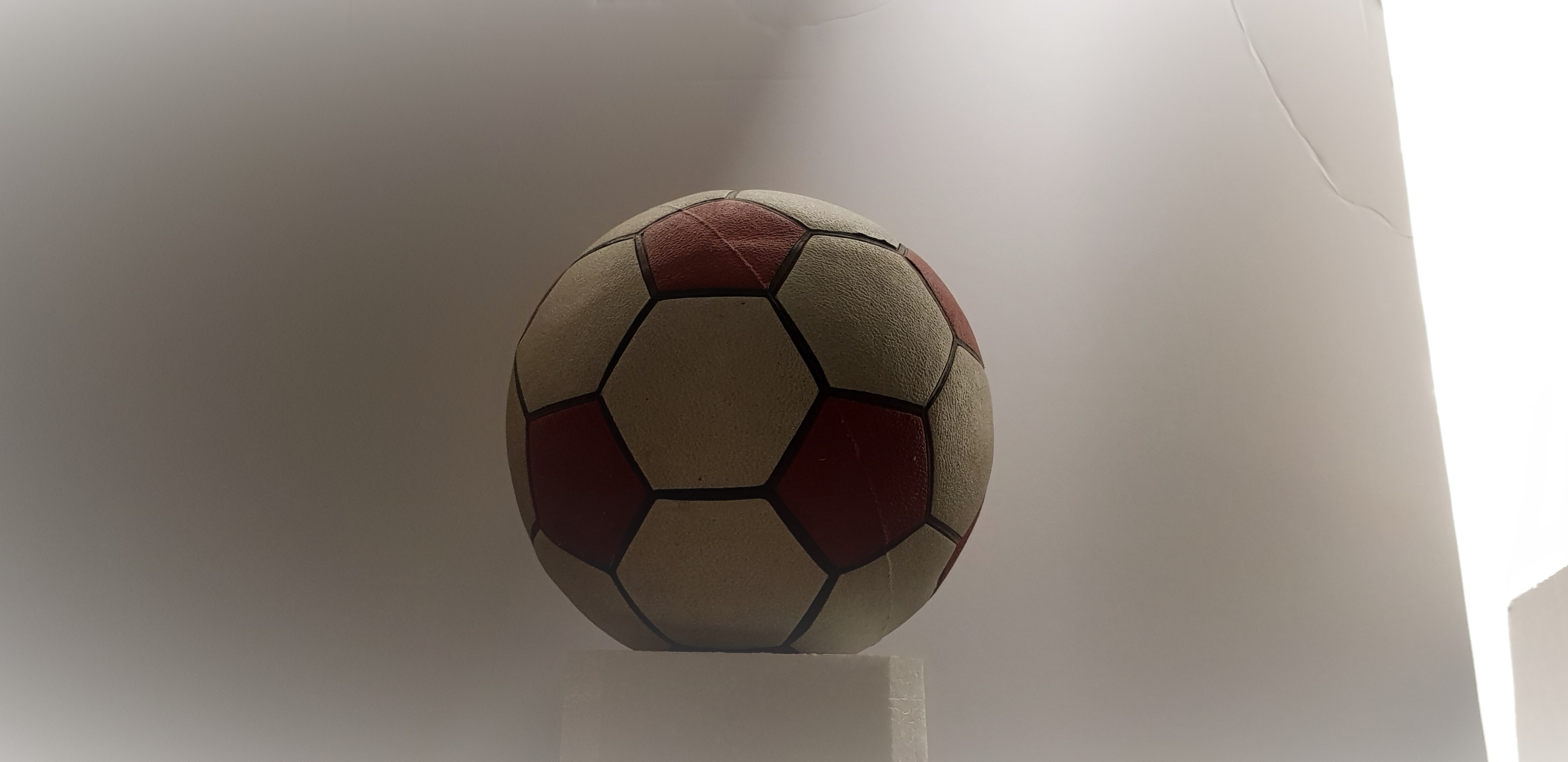}   
   \caption{Scene}
   \label{fig:Jc1} 
\end{subfigure}\vspace{10mm}

 \begin{subfigure}[b]{\linewidth} 
\centering
   \includegraphics[scale=1.5]{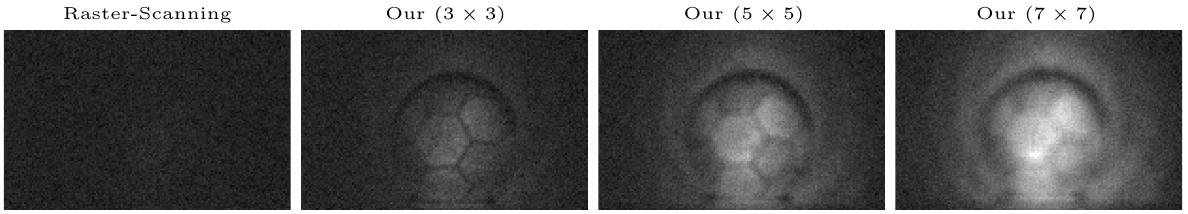}       
   \caption{Experimental data. A display of observation vectors $ \bm {\vec v}$, Eq.~\ref {eq:Multi}.
	\textit{Supplementary details:} for a consistent comparison, images shown here have the same photon-count to grayscale map (i.e., a black pixel represents the minimum photon-count among all the four images. Similarly, a white pixel represents the maximum photon-count among all the four images. Photon-counts between the maximum and minimum are mapped to a gray color). For the proposed scheme, the illumination window size ($w\times w$) is varied from $3\times 3$ to $ 7\times 7$. All images are gamma-corrected, $\gamma= 2.2$.}
   \label{fig:Rv}
\end{subfigure}\vspace{12mm}

 \begin{subfigure}[b]{\linewidth} 
\centering
   \includegraphics[scale=1.5]{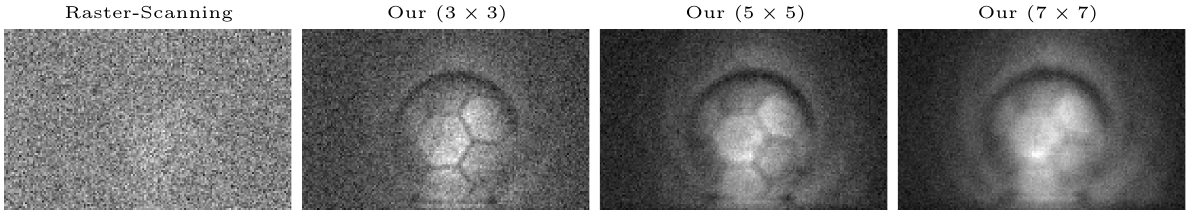}       
   \caption{Experimental data. Images of Fig.~\ref{fig:Rv} with an independent grayscale map.
	\textit{Supplementary details:} to enhance, to some extent, the visibility of the images shown  in Fig.~\ref{fig:Rv}, each  image here has an independent grayscale map (i.e., the minimum and maximum photon-counts of each individual image is mapped to a black and white pixel, respectively. Photon-counts between the maximum and minimum are  mapped to a gray color). All images are gamma-corrected, $\gamma= 2.2$.}
   \label{fig:Jc3}
\end{subfigure}\vspace{12mm}

 \begin{subfigure}[b]{\linewidth} 
\centering
	\includegraphics[scale=1.5]{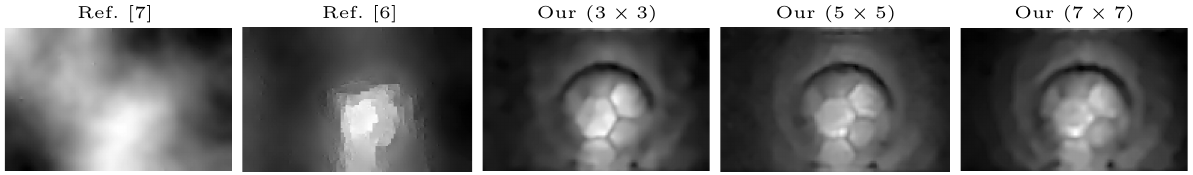}  

\caption{The end-results. Denoised/deconvolved images.
 \textit{Supplementary details:} the input data for both Ref.~\cite{System}  and \cite{First_Photon} is obtained from the image Raster-Scanning  displayed in Fig.~\ref{fig:Jc3}. For Ref.~\cite{First_Photon}, the input data is the number of  transmitted pulses until the first photon arrives. For Ref.~\cite{System}, the input noisy image is the intensity image labeled Raster-Scanning in Fig.~\ref{fig:Jc3}. For the proposed scheme, the input noisy/blurred images for $3 \times 3$, $5 \times 5$, and $7 \times 7$,  are their corresponding images shown in Fig.~\ref{fig:Jc3}. All images here have an independent grayscale map and are gamma-corrected, $\gamma= 2.2$.}

\label{fig:AllD}
\end{subfigure}
\vspace{7mm}
\caption{Intensity images.}\label{fig:Ia}

\end{figure*}

\begin{figure*}[!htb]
    \centering
	    \begin{subfigure}[b]{0.2\textwidth}
 \includegraphics[scale=.18]{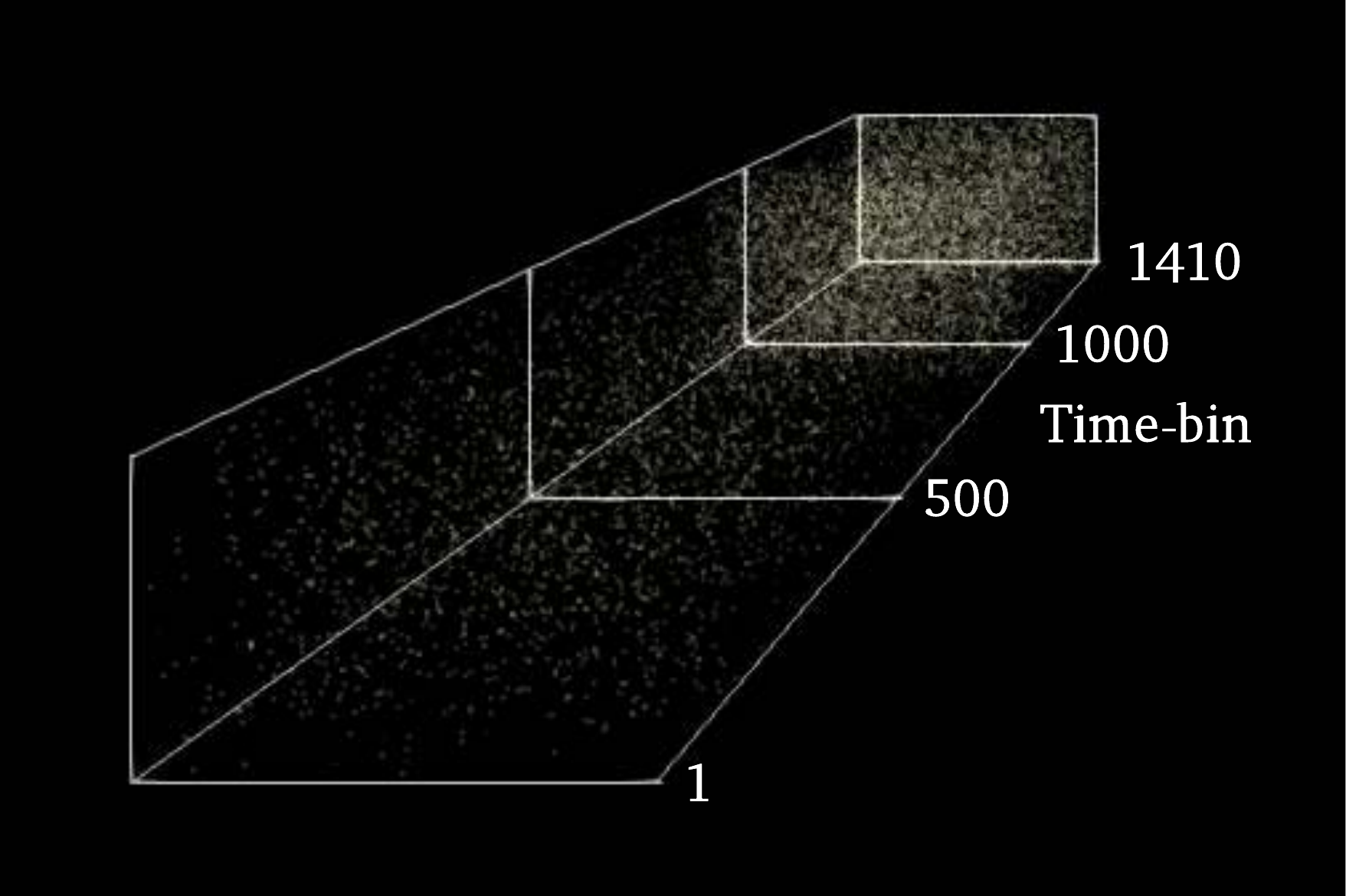}        
 \caption{\small{First photon of~~~~~\\~~~~~raster-scanning}}
        \label{fig:fPD}
    \end{subfigure}%
    \begin{subfigure}[b]{0.2\textwidth}
        \includegraphics[scale=.177]{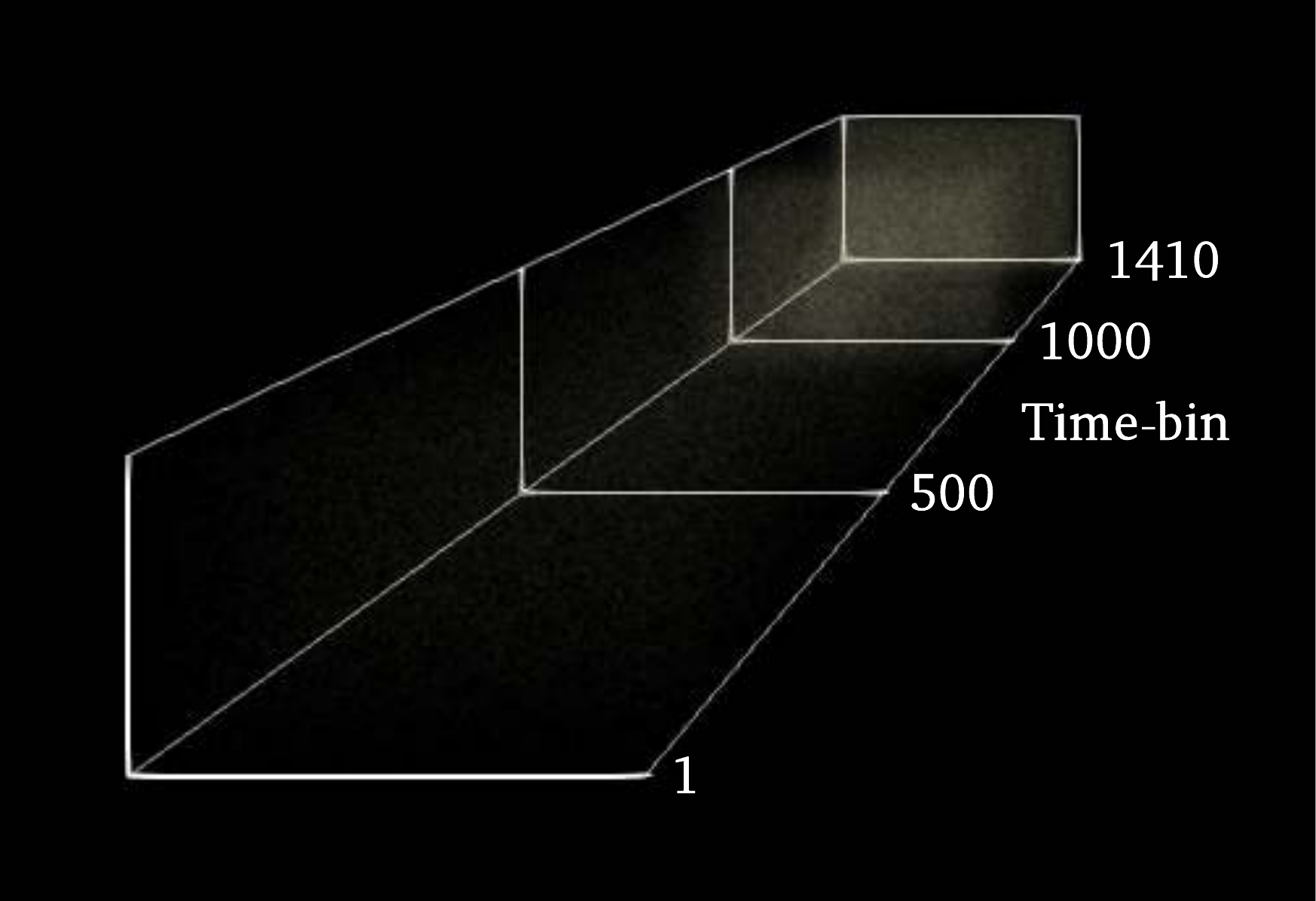}
\caption{\small{All photons of \\raster-scanning}}
        \label{fig:rD}
    \end{subfigure}%
		  \begin{subfigure}[b]{0.22\textwidth}
        \includegraphics[scale=.188]{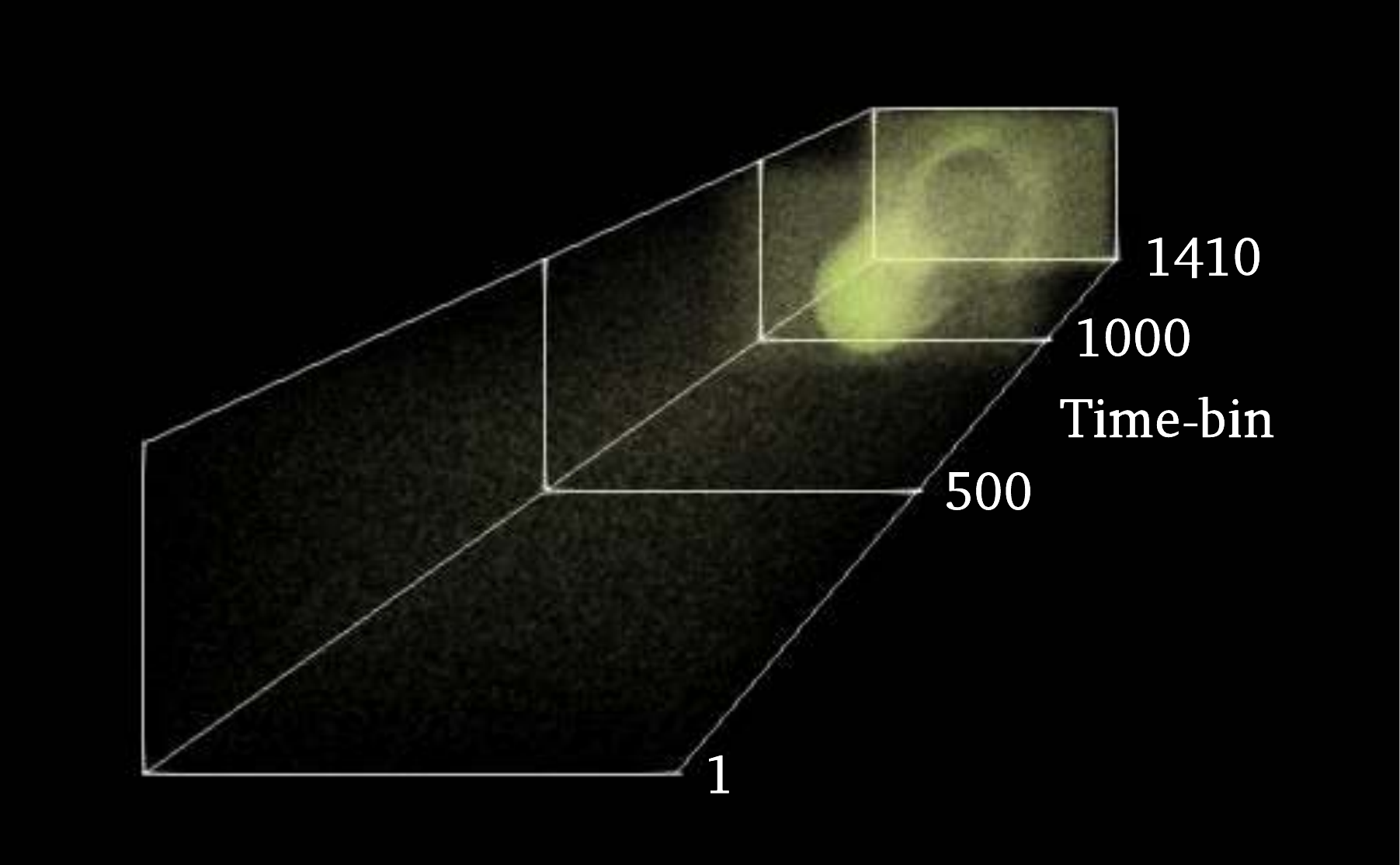}
   \caption{Our $(3\times 3)$\\~~~}
        \label{fig:3D}
    \end{subfigure}%
		  \begin{subfigure}[b]{0.2\textwidth}
        \includegraphics[scale=.175]{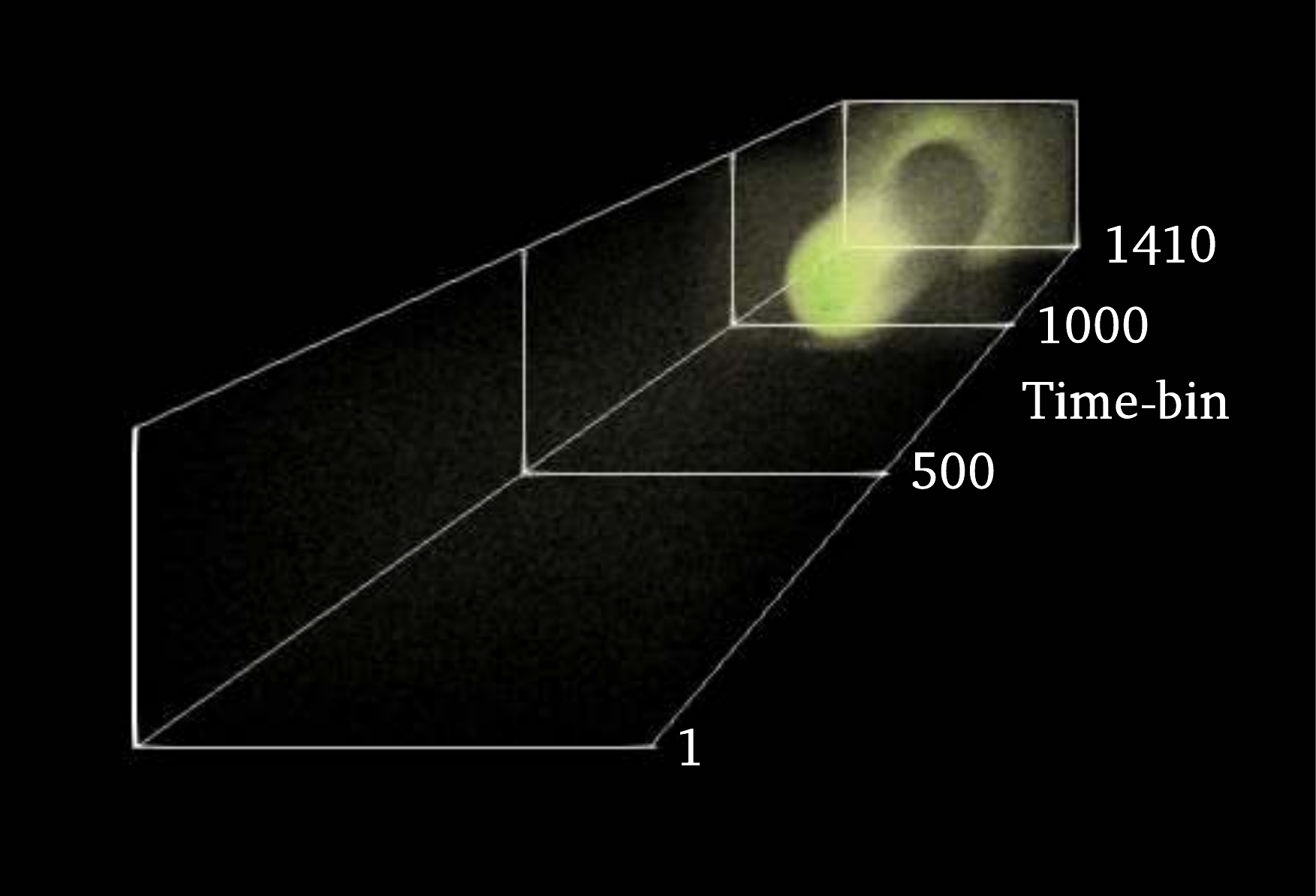}
     \caption{Our $(5\times 5)$\\~~~}
        \label{fig:5D}
    \end{subfigure}%
				  \begin{subfigure}[b]{0.19\textwidth}
        \includegraphics[scale=.181]{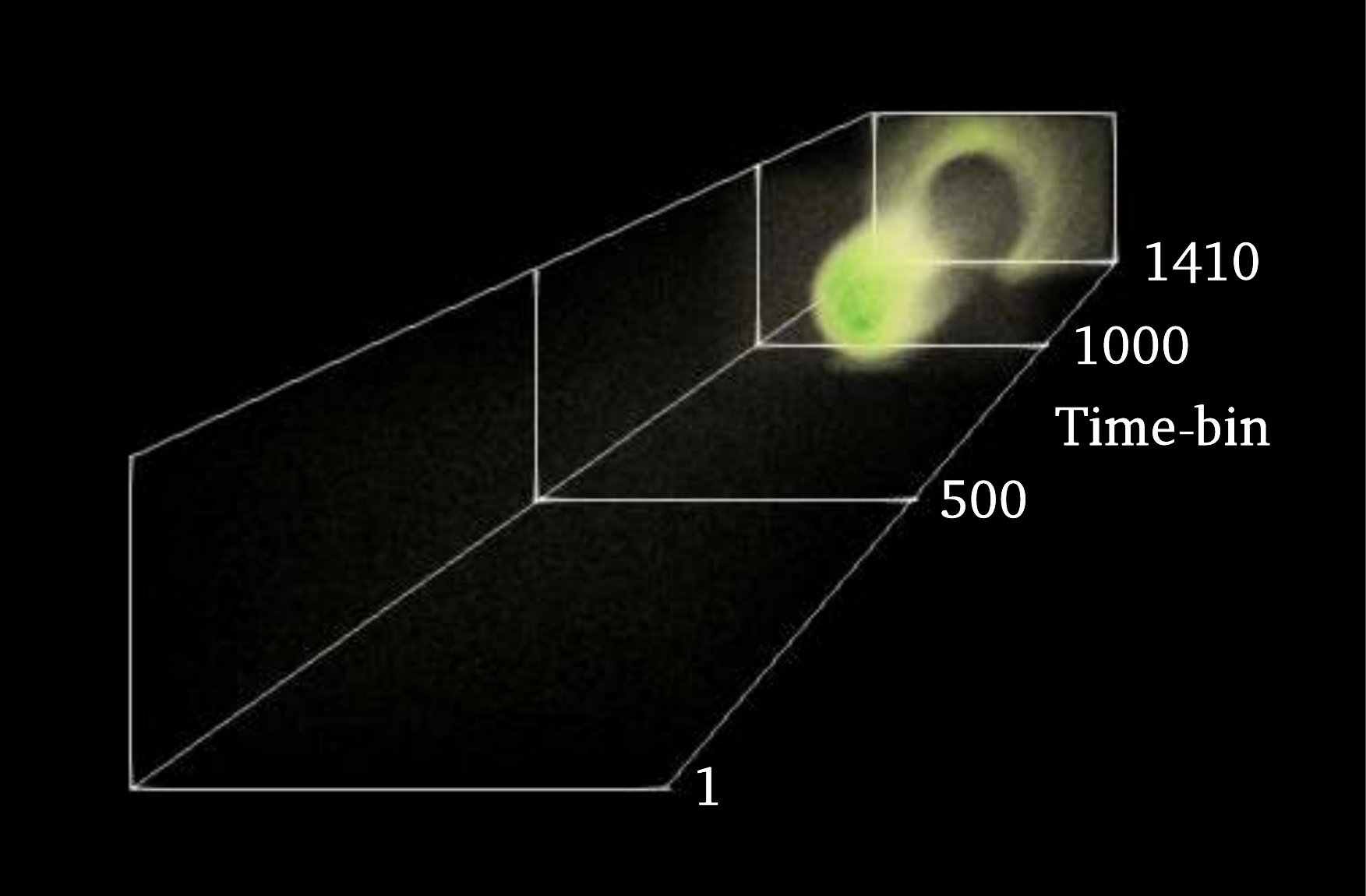}
        \caption{Our $(7\times 7)$\\~~~}
        \label{fig:7D}
    \end{subfigure}\\[3ex]
		
    ~ 
    \begin{subfigure}[b]{0.1975\textwidth}
        \includegraphics[scale=.2]{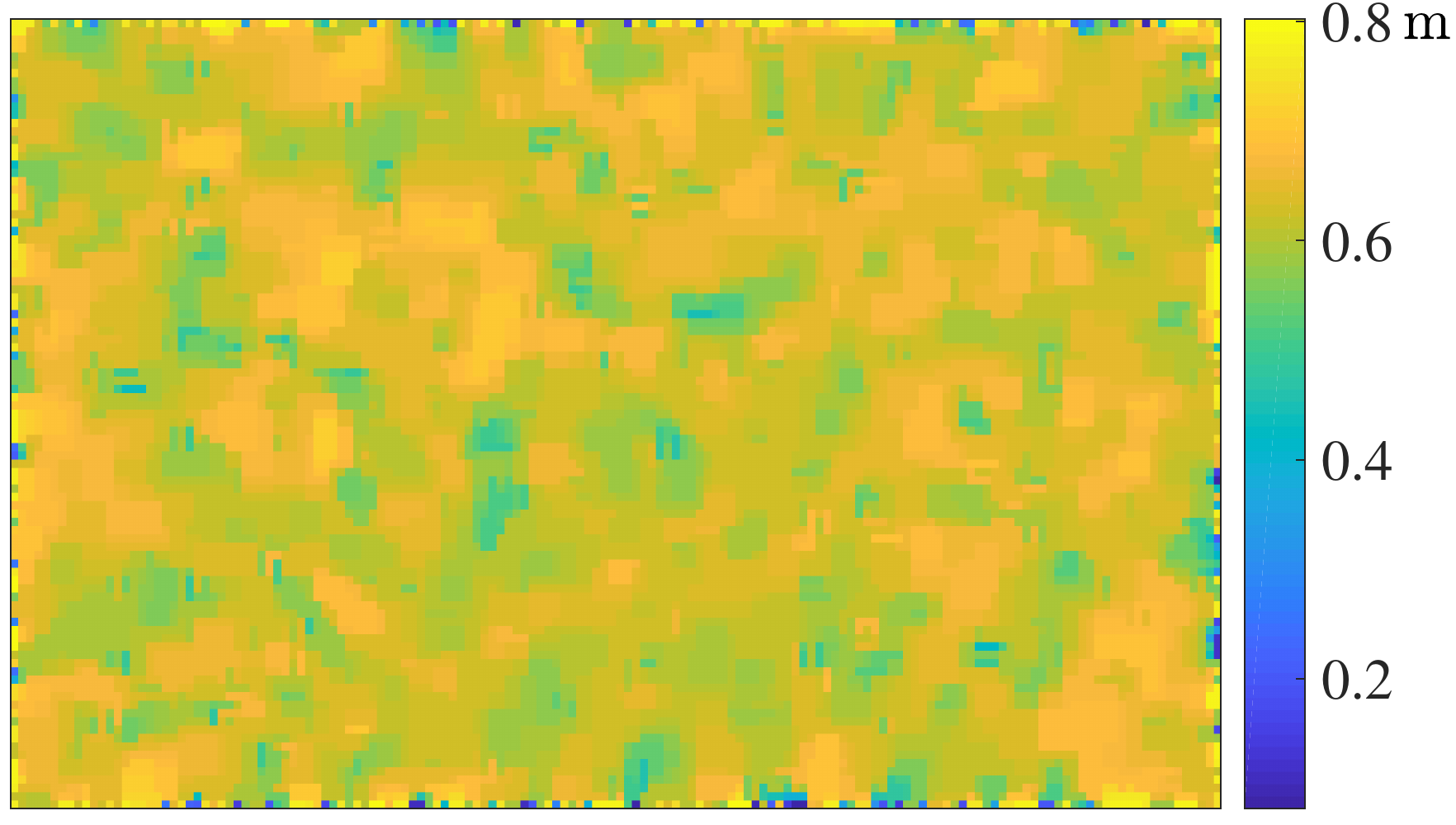}
         \caption{Ref.~[7]}
        \label{fig:First3}
    \end{subfigure}%
		    \begin{subfigure}[b]{0.201\textwidth}
        \includegraphics[scale=.2]{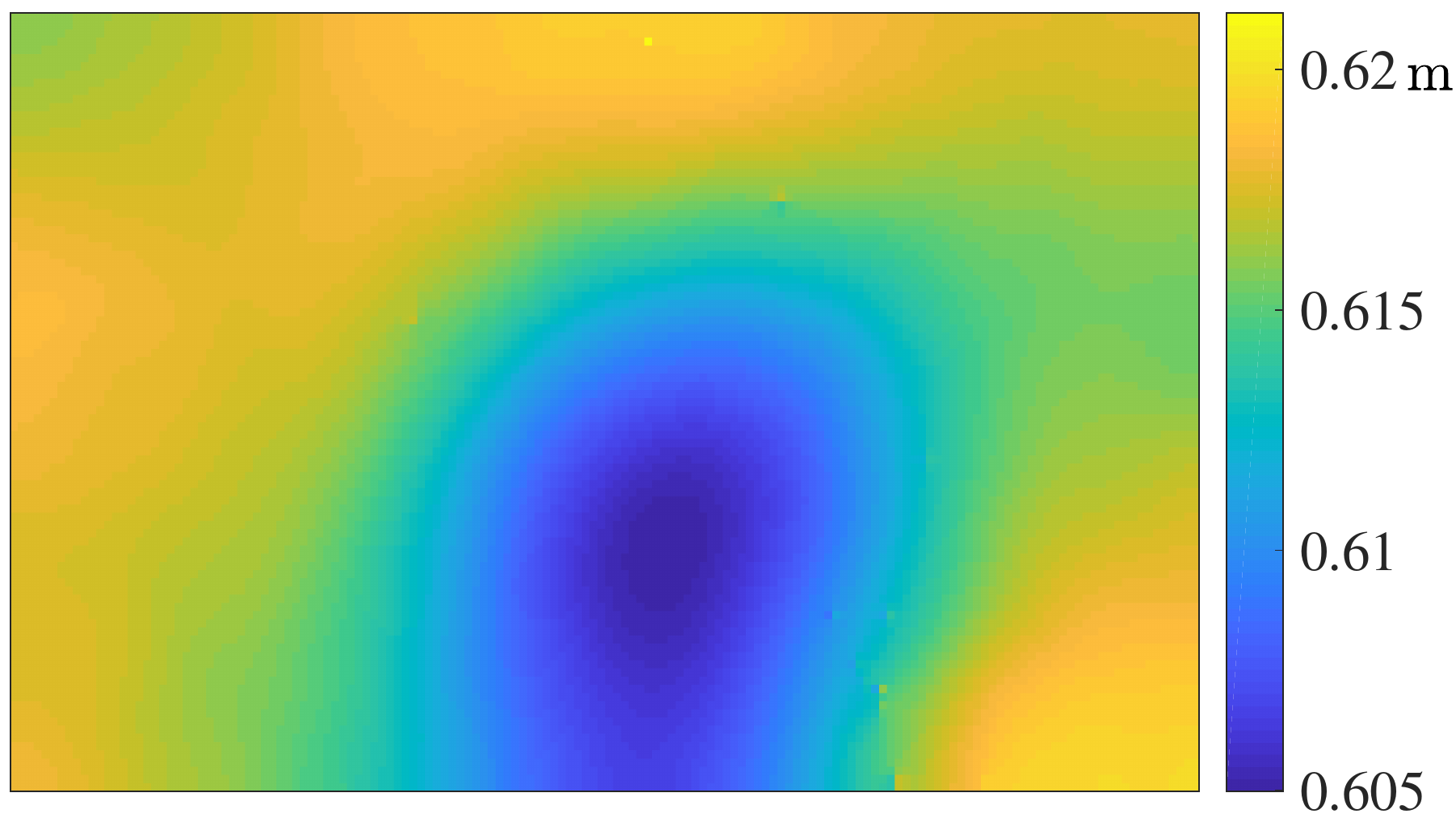}
        \caption{Ref.~[6]}
        \label{fig:Shin3}
    \end{subfigure}%
				    \begin{subfigure}[b]{0.201\textwidth}
        \includegraphics[scale=.205]{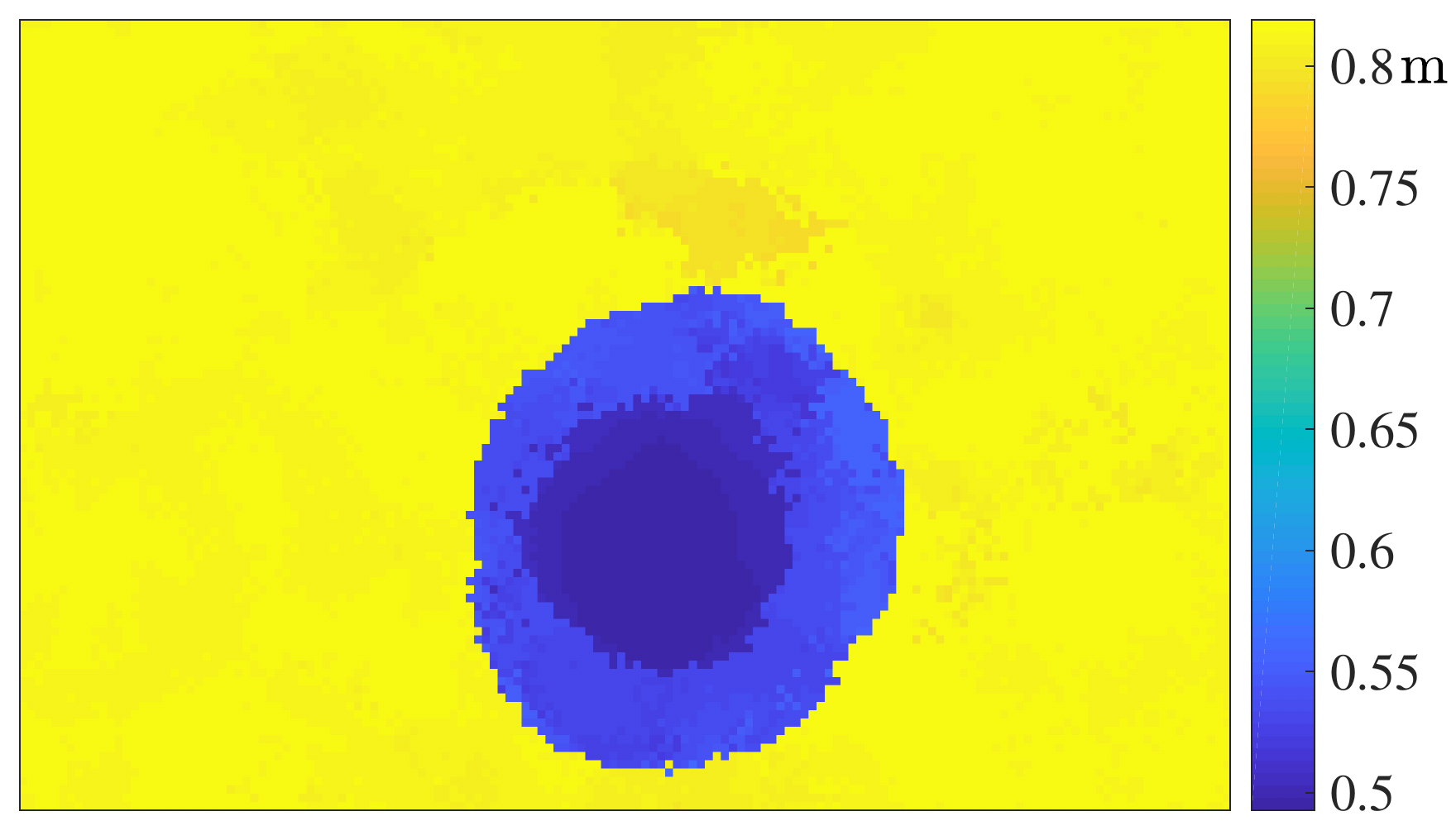}
        \caption{Our $(3\times 3)$}
        \label{fig:3d}
    \end{subfigure}%
			    \begin{subfigure}[b]{0.197\textwidth}
        \includegraphics[scale=.2]{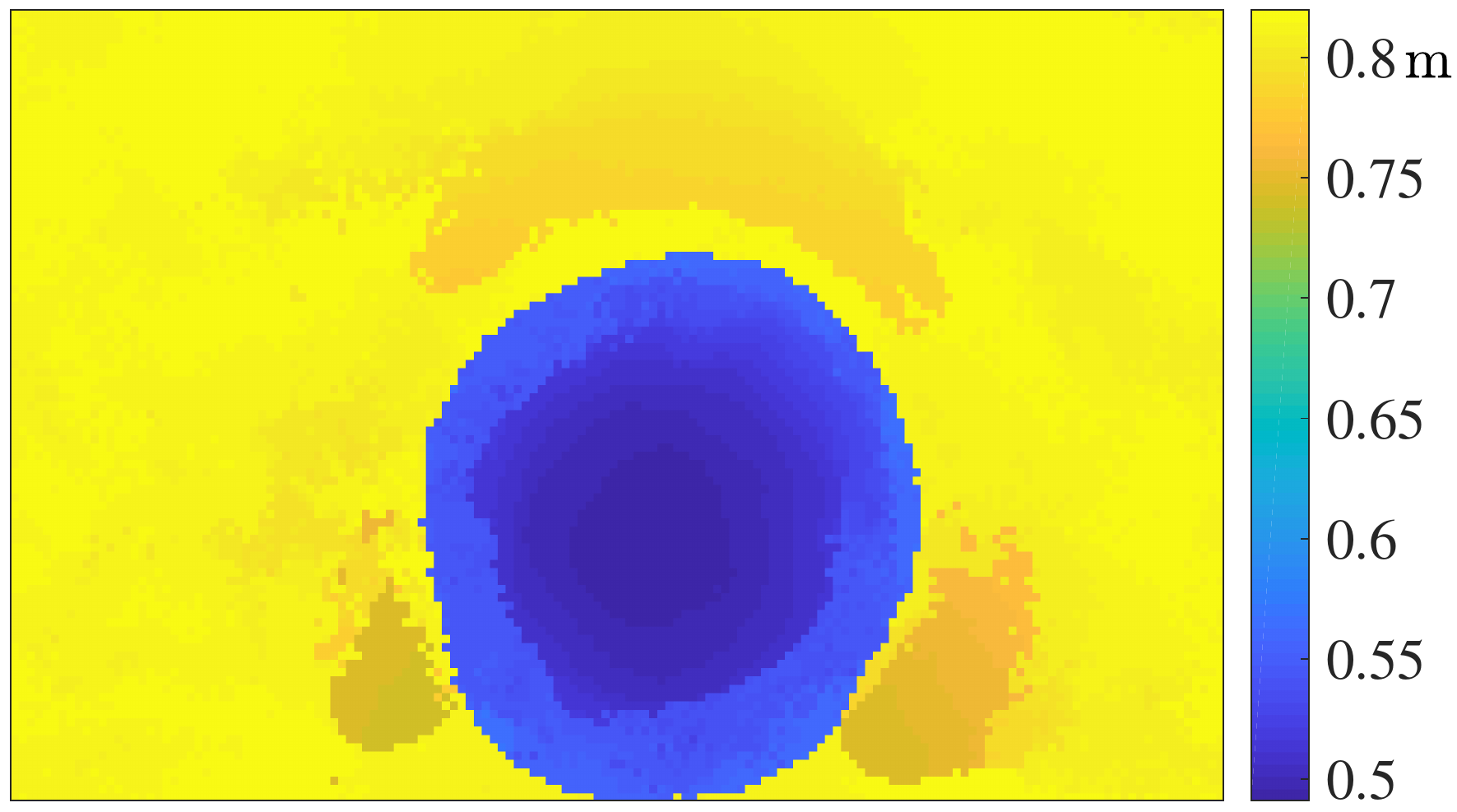}
     \caption{Our $(5\times 5)$}
        \label{fig:5d}
    \end{subfigure}%
			    \begin{subfigure}[b]{0.205\textwidth}
        \includegraphics[scale=.2]{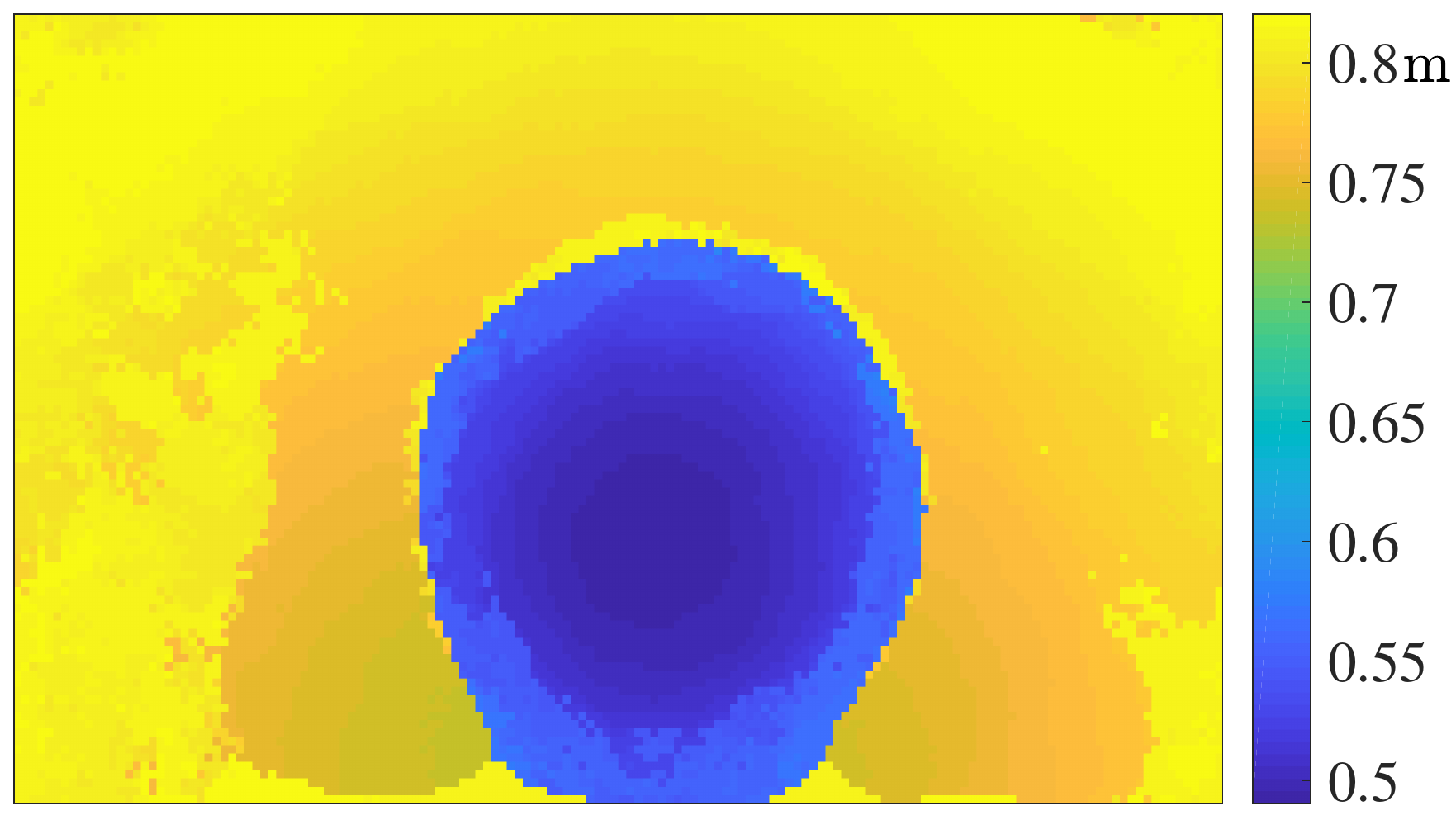}
        \caption{Our $(7\times 7)$}
        \label{fig:7d}
    \end{subfigure}
    \caption{ (a)-(e) Raw data: a visualization of experimental data (first photon of raster-scanning and matrix $\bm R$, Eq.~\ref{eq:multiM5}). (f)-(j) Depth images.
		\textit{Supplementary details}: each column of $\bm R$ is reshaped to form a $95 \times 152$ image, so as to visualize $\bm R$ in a 3D space ($95 \times 152 \times 1410$). In the experiment,  we vary the illumination window size from $3\times 3$ to $ 7\times 7$ for the proposed scheme. The opacity of points in figure  Fig.~\ref{fig:fPD} are considerably higher than that of figures in the first row, so as to enhance visibility.}
    \label{fig:3Dr}
\end{figure*}

Fig.~\ref{fig:Rv} and Figs.~\ref{fig:fPD}--\ref{fig:7D}, display data gathered from the experiment for the observation vector $ \bm {\vec v}$ (Eq.~\ref {eq:Multi}) and matrix $\bm R$ (Eq.~\ref{eq:multiM5}), respectively. Fig.~\ref{fig:AllD} and Figs.~\ref{fig:3d}--\ref{fig:7d} present  the performance of the proposed scheme.

A rich body of research exists on  3D structure and reflectivity imaging with photon-counting detectors. However, this is the first paper to address the issue of diffraction-photons in such systems. We therefore believe that, in addition to presenting the performance of our scheme, it would be beneficial to compare\;it\;with commonly\;used\;benchmarks\,(\!\!\cite{System}~and~\cite{First_Photon}).

Fig.~\ref{fig:Ia} and \ref{fig:3Dr} show  that the proposed scheme reveals a latent image amid a bath of  diffraction-photons ($\sim$\,$25$ photons/pixel, Fig.~\ref{Fig:Histograms_All}). Methods based on traditional raster-scanning \cite{First_Photon,System}, however, are adversely affected by diffraction-photons.  The reason for this is that the average number of signal photons per pixel for raster-scanning is greatly below the number of diffraction-photons; this results in diminished image quality.

The proposed scheme has an improved performance as it takes co-designed illumination and deconvolution approach to solve the image capturing problem, which boosts the average number of signal photons collected per pixel (Fig.~\ref{Fig:Histograms_All}) and maintains the native image resolution  as the relatively large illumination blocks overlap (Fig.~\ref {Fig:Illumination}). For our experiment, an illumination window size of $5 \times 5$ produced the best results: this size is a workable balance between signal photons collected and blur introduced.

It is important to point out, however, that the performance improvement of the proposed scheme over~\cite{System} and~\cite{First_Photon} should be viewed with some caution, because~\cite{System} and~\cite{First_Photon}  are designed for imaging environments without diffraction-photons. As such, they produce a lower image quality than our scheme.

\section{Conclusion}
\label{sec:c}

This study set out to overcome diffraction-photons. We  described  a method  for acquiring intensity and depth images in an environment tainted by diffraction-photons ($\sim 25$ photons/pixel). 

A proof-of-concept experiment demonstrates the viability of the designed scheme. It is shown that at a low DMD contrast ratio ($\sim$\,$1000{:}1$), diffraction-photons can be overcome via a joint illumination-deconvolution scheme, enabling the acquisition of intensity and depth images.

The scheme works as the number of signal photons collected is boosted by overlapping sizeable illumination blocks. The overlapping blocks mix pixel values, which are subsequently untangled by deconvolution algorithms.

The central conclusion that can be drawn from this study is that the designed scheme offers a means to relax the trade-off between spatial resolution and signal power---this is achieved mainly through convex optimization techniques.

A promising avenue for future research is to determine the optimal illumination window size mathematically,  for a given  contrast ratio, deadtime and noise level. Thus far, the optimal window size has been determined by experimental testing. Additionally, a potential research direction is to analyze the sensitivity of the problems in Eqs.~\ref{eq:Mnop} and~\ref{eq:CT} to noise in $\bm H$. Such an analysis would provide insight into how to best deconvolve images captured by SPAD/DMD-based imaging systems.

To conclude,  we believe that the findings in this paper will be of special interest to researchers in the field of ToF imaging as it addresses a new practical challenge.


\appendices

\section{alternating direction method {of} multipliers} \label{sec:App1}

In this section we shall denote the shrinkage operator (element-wise soft thresholding) by

$$
\mathcal{S}_{\tau} (x) \defeq
\begin{cases}
     x-\tau,& x>\tau  \\
       0,& |x|\leq\tau    \\
			x+\tau,& x<-\tau 
\end{cases}
$$

\noindent Additionally, we denote the projection of $x$ onto set $\Omega=\left[ 2\sqrt{3/8} , +\infty\right)$ and $\mathbb R_+$ by $\text{Proj}_{\Omega}(x)\defeq\max\left\{x,2\sqrt{3/8}\right\}$ and $\text{Proj}_{\mathbb R_+}(x)\defeq\max\{x,0\}$, respectively.

\begin{algorithm}[!htb]
\nonl Initialize $\bm z^{(0)}_1$, $\bm z^{(0)}_2$, $\bm u^{(0)}_1$,  $\bm u^{(0)}_2$,  \text{set} $k=0$ and \\
\nonl choose $\rho_1$, $\rho_2$, $\mu>0$\\
	\textbf{repeat} \\
	\medskip
	\nonl~~~~~~~~~\textcolor{gray}{//$b$-minimization}\\
 ~~~~~~~~~$\bm{ \vec b}^{(k+1)}  \leftarrow \argminl_{\vec {\bm b}} \quad \dfrac{1}{2} \lVert  \vec {\bm b}- f(\bm{\vec v})\lVert^2_2$\\
 \nonl~\hspace{3.3cm}$\quad +\quad\dfrac{\rho_1}{2}  \Vert \bm D \vec {\bm b} - {\bm z^{(k)}_1}+\bm u_1^{(k)}\Vert^2_2$\\ 
	\smallskip
\nonl ~\hspace{3.3cm}$\quad +\quad\dfrac{\rho_2}{2} \lVert\vec {\bm b}- {\bm z^{(k)}_2} +\bm u_2^{(k)}\lVert^2_2$\\
	\medskip
	\nonl~~~~~~~\textcolor{gray}{//$z$-minimizations}\\
	
	~~~~~~~~~$\bm z^{(k+1)}_1 \leftarrow \mathcal{S}_{\mu/\rho_1} \left(\bm D  {\bm{ \vec b}}^{(k+1)} +\bm u^{(k)}_1\right)$\\
		\smallskip
~~~~~~~~~$\bm z^{(k+1)}_2 \leftarrow \text{Proj}_{\Omega} \left( {\bm{ \vec b}}^{(k+1)} +\bm u^{(k)}_2\right)$\\
	\medskip
	\nonl~~~~~~~~~\textcolor{gray}{//dual updates}\\	
 ~~~~~~~~~$\bm u^{(k+1)}_1 \leftarrow \bm u^{(k)}_1 +\bm D {\bm{ \vec b}}^{(k+1)} -\bm z^{(k+1)}_1$\\
			\smallskip
 ~~~~~~~~~$\bm u^{(k+1)}_2 \leftarrow \bm u^{(k)}_2 +{\bm{\vec b}}^{(k+1)} -\bm z^{(k+1)}_2$\\
\medskip
	\nonl~~~~~~~~~\textcolor{gray}{//}\\
 ~~~~~~~~~$k\leftarrow k+1$\\
	\medskip
	
\textbf{until} stopping criteria is satisfied.

	\caption{ADMM algorithm for minimizing Eq.~\ref{eq:De}}
\label{algo:Admm1}
	\end{algorithm}

\begin{algorithm}[!htb]
\nonl Initialize $\bm z^{(0)}_1$, $\bm z^{(0)}_2$, $\bm u^{(0)}_1$,  $\bm u^{(0)}_2$,  set $k=0$ and\\
 \nonl choose $\rho_1$, $\rho_2$, $\lambda>0$\\
	\textbf{repeat} \\
	\medskip
	\nonl~~~~~~~~~\textcolor{gray}{//$\alpha$-minimization}\\
~~~~~~~~~$\bm {\vec \alpha}^{(k+1)} \leftarrow \argminl_{\bm {\vec \alpha}} \quad \dfrac{1}{2} \lVert  \bm H\bm {\vec \alpha}- \vec {\bm b}_{\tiny *}\lVert^2_2$\\
 \nonl~\hspace{3.4cm}$\quad +\quad\dfrac{\rho_1}{2}  \Vert \bm D \bm{\vec \alpha} - {\bm z^{(k)}_1} +\bm u_1^{(k)}\Vert^2_2$\\
\smallskip
  \nonl~\hspace{3.4cm}$\quad+\quad \dfrac{\rho_2}{2} \lVert \bm {\vec \alpha} - {\bm z^{(k)}_2} +\bm u_2^{(k)}\lVert^2_2$\\
	\medskip
	\nonl~~~~~~~~~\textcolor{gray}{//$z$-minimizations}\\
	
~~~~~~~~~$\bm z^{(k+1)}_1 \leftarrow \mathcal{S}_{\lambda/\rho_1} \left(\bm D \bm {\vec \alpha}^{(k+1)} +\bm u^{(k)}_1\right)$\\
		\smallskip
~~~~~~~~~$\bm z^{(k+1)}_2 \leftarrow \text{Proj}_{\mathbb R_+} \left( \bm {\vec \alpha}^{(k+1)} +\bm u^{(k)}_2\right)$\\	
	\medskip
	\nonl~~~~~~~~~\textcolor{gray}{//dual updates}\\	
~~~~~~~~~$\bm u^{(k+1)}_1 \leftarrow \bm u^{(k)}_1 +\bm D\bm {\vec \alpha}^{(k+1)} -\bm z^{(k+1)}_1$\\
			\smallskip
~~~~~~~~~$\bm u^{(k+1)}_2 \leftarrow \bm u^{(k)}_2 +\bm {\vec \alpha}^{(k+1)} -\bm z^{(k+1)}_2$\\
\medskip
\nonl~~~~~~~~~\textcolor{gray}{//}\\
~~~~~~~~~$k\leftarrow k+1$\\
	\medskip
\textbf{until} stopping criteria is satisfied.

	\caption{ADMM algorithm for minimizing Eq.~\ref{eq:Mnop}}
\label{algo:Admm2}
	\end{algorithm}

\begin{algorithm}[htb!]
\nonl Initialize $\bm z^{(0)}_1$, $\bm z^{(0)}_2$, $\bm u^{(0)}_1$,  $\bm u^{(0)}_2$,  set $k=0$ and\\
 \nonl choose $\rho_1$, $\rho_2$, $\mu>0$ \\
	\textbf{repeat} \\
	\medskip
	\nonl~~~~~~~~~\textcolor{gray}{//$C$-minimization}\\
~~~~~~~~~$ C_j^{(k+1)} \leftarrow \argminl_{C} \quad \dfrac{1}{2} \lVert  \bm H C_j- R_j\lVert^2_2$\\
\nonl~\hspace{3.39cm}$\quad +\quad\dfrac{\rho_1}{2}  \Vert \nabla C_j - {\bm z^{(k)}_1} +\bm u_1^{(k)}\Vert^2_2$\\
\smallskip
\nonl~\hspace{3.39cm}$ \quad+\quad \dfrac{\rho_2}{2} \lVert C_j - {\bm z^{(k)}_2} +\bm u_2^{(k)}\lVert^2_2$\\
	\medskip
	\nonl~~~~~~~~~\textcolor{gray}{//$z$-minimizations}\\
	
~~~~~~~~~$\bm z^{(k+1)}_1 \leftarrow \mathcal{S}_{\mu/\rho_1} \left(\nabla {C_j}^{(k+1)} +\bm u^{(k)}_1\right)$\\
		\smallskip
~~~~~~~~~$\bm z^{(k+1)}_2 \leftarrow \text{Proj}_{\mathbb R_+} \left( {C_j}^{(k+1)} +\bm u^{(k)}_2\right)$\\	
	\medskip
	\nonl~~~~~~~~~\textcolor{gray}{//dual updates}\\	
~~~~~~~~~$\bm u^{(k+1)}_1 \leftarrow \bm u^{(k)}_1 +\nabla{C_j}^{(k+1)} -\bm z^{(k+1)}_1$\\
			\smallskip
~~~~~~~~~$\bm u^{(k+1)}_2 \leftarrow \bm u^{(k)}_2 +{C_j}^{(k+1)} -\bm z^{(k+1)}_2$\\
\medskip
	\nonl~~~~~~~~~\textcolor{gray}{//}\\
~~~~~~~~~$k\leftarrow k+1$\\
	\medskip
\textbf{until} stopping criteria is satisfied.

	\caption{ADMM algorithm for minimizing Eq.~\ref{eq:CT}}
\label{algo:Admm3}
\end{algorithm}

\newpage

\section{Derivative Matrices}\label{sec:App2}

In this section we describe the derivative matrices used throughout the paper. Let $\bm D\in \mathbb R^{8n\times n}$ be defined as follows:

\begingroup
\begin{equation*} \label{eq:Dmatrix}
  \bm D=\begin{pmatrix} 
  \nabla\\
   ~\rho\nabla^{''}\\
  \end{pmatrix}_{\!\!8n \times n}
\end{equation*}
\endgroup

\noindent where $\nabla$ and $\nabla^{''}$ are the first and second order derivative matrices, respectively;  and constant $\rho$ controls the strength of the second derivative. Here, the first  and second derivative matrices are defined as

\begingroup
\renewcommand*{\arraystretch}{1.3}
\begin{equation*} \label{eq:D1matrix}
  \nabla=\begin{pmatrix} 
  \mathcal{A}_x\\
	\mathcal{A}_y\\
	\mathcal{A}_{xy}\\
	\mathcal{A}_{yx}
  \end{pmatrix}_{\!\!4n \times n}
\end{equation*}
\endgroup
\noindent and

\begingroup
\renewcommand*{\arraystretch}{1.4}
\begin{equation*} \label{eq:D1matrix}
\nabla^{''}=\begin{pmatrix} 
 \mathcal{A}^{''}_x\\
	\mathcal{A}^{''}_y\\
	\mathcal{A}^{''}_{xy}\\
	\mathcal{A}^{''}_{yx}
  \end{pmatrix}_{\!\!4n \times n}
\end{equation*}
\endgroup

\noindent where $ \mathcal{A},~  \mathcal{A}^{''}\in\mathbb R^{n\times n}$ are convolution matrices for the first and second  derivatives, respectively; and their subscripts specify the direction in which a  derivative operation is performed.

The rationale of including a  second derivative in our regularizer is that it encourages the recovery of image curvatures, rendering deblurred images more naturally-looking.


\section*{Acknowledgment}
The authors wish to express their gratitude to Congli Wang, Muxingzi Li and Qilin Sun for their technical assistance and fruitful discussions. This research was made possible by KAUST Baseline Funding.

\nocite{Harmany}

\balance
\bibliographystyle{IEEEtran}
\bibliography{MyB}

\begin{thebibliography}{10}
\providecommand{\url}[1]{#1}
\csname url@samestyle\endcsname
\providecommand{\newblock}{\relax}
\providecommand{\bibinfo}[2]{#2}
\providecommand{\BIBentrySTDinterwordspacing}{\spaceskip=0pt\relax}
\providecommand{\BIBentryALTinterwordstretchfactor}{4}
\providecommand{\BIBentryALTinterwordspacing}{\spaceskip=\fontdimen2\font plus
\BIBentryALTinterwordstretchfactor\fontdimen3\font minus
  \fontdimen4\font\relax}
\providecommand{\BIBforeignlanguage}[2]{{%
\expandafter\ifx\csname l@#1\endcsname\relax
\typeout{** WARNING: IEEEtran.bst: No hyphenation pattern has been}%
\typeout{** loaded for the language `#1'. Using the pattern for}%
\typeout{** the default language instead.}%
\else
\language=\csname l@#1\endcsname
\fi
#2}}
\providecommand{\BIBdecl}{\relax}
\BIBdecl

\bibitem{Gariepy}
G.~Gariepy, N.~Krstaji{\'c}, R.~Henderson, C.~Li, R.~R. Thomson, G.~S. Buller,
  B.~Heshmat, R.~Raskar, J.~Leach, and D.~Faccio, ``{Single-Photon Sensitive
  Light-in-F[l]ight Imaging},'' \emph{Nature communications}, vol.~6, p. 6021,
  2015.

\bibitem{NLOS}
M.~O'Toole, D.~B. Lindell, and G.~Wetzstein, ``{Confocal Non-Line-of-Sight
  Imaging Based on the Light-Cone Transform},'' \emph{Nature}, 2018.

\bibitem{MPD}
{PicoQuant}, ``{PDM Series - Single Photon Avalanche Diodes},''
  \emph{Datasheet}, {September 2015}.

\bibitem{TI}
{Texas Instruments}, ``{DLP4500 (0.45 WXGA DMD)},'' \emph{DLPS028C datasheet},
  {April 2013 [Revised Feb. 2018]}.

\bibitem{Handbook}
P.~Hinterdorfer and A.~Van~Oijen, \emph{{Handbook of Single-Molecule
  Biophysics}}.\hskip 1em plus 0.5em minus 0.4em\relax Springer Science \&
  Business Media, {2009, pp. 73}.

\bibitem{System}
D.~Shin, A.~Kirmani, V.~K. Goyal, and J.~H. Shapiro, ``{Photon-Efficient
  Computational 3-D and Reflectivity Imaging with Single-Photon Detectors},''
  \emph{IEEE Transactions on Computational Imaging}, vol.~1, no.~2, pp.
  112--125, 2015.

\bibitem{First_Photon}
A.~Kirmani, D.~Venkatraman, D.~Shin, A.~Cola{\c{c}}o, F.~N. Wong, J.~H.
  Shapiro, and V.~K. Goyal, ``{First-Photon Imaging},'' \emph{Science}, vol.
  343, no. 6166, pp. 58--61, 2014.

\bibitem{Stanford}
M.~O'Toole, F.~Heide, D.~B. Lindell, K.~Zang, S.~Diamond, and G.~Wetzstein,
  ``{Reconstructing Transient Images from Single-Photon Sensors},'' in
  \emph{Proc. IEEE CVPR}, 2017, pp. 2289--2297.

\bibitem{Photonics}
W.~He, Z.~Feng, J.~Lin, S.~Shen, Q.~Chen, G.~Gu, B.~Zhou, and P.~Zhang,
  ``{Adaptive Depth Imaging with Single-Photon Detectors},'' \emph{IEEE
  Photonics Journal}, vol.~9, no.~2, pp. 1--12, 2017.

\bibitem{Shin}
D.~Shin, F.~Xu, D.~Venkatraman, R.~Lussana, F.~Villa, F.~Zappa, V.~K. Goyal,
  F.~N. Wong, and J.~H. Shapiro, ``{Photon-Efficient Imaging with a
  Single-Photon Camera},'' \emph{Nature communications}, vol.~7, p. 12046,
  2016.

\bibitem{ThorlabsD}
{Thorlabs, Inc}, ``{GVS011 and GVS012 Large Mirror Diameter Scanning Galvo
  Systems},'' \emph{User Guide}, {HA0220T [Revised 7 July 2011]}.

\bibitem{Sweep2}
G.~A. Howland, P.~Zerom, R.~W. Boyd, and J.~C. Howell, ``{Compressive Sensing
  LIDAR for 3D Imaging},'' in \emph{Lasers and Electro-Optics (CLEO), 2011
  Conference on}.\hskip 1em plus 0.5em minus 0.4em\relax IEEE, 2011, pp. 1--2.

\bibitem{Sweep3}
L.~Li, L.~Wu, X.~Wang, and E.~Dang, ``{Gated Viewing Laser Imaging with
  Compressive Sensing},'' \emph{Applied optics}, vol.~51, no.~14, pp.
  2706--2712, 2012.

\bibitem{Sweep4}
L.~Li, W.~Xiao, and W.~Jian, ``{Three-Dimensional Imaging Reconstruction
  Algorithm of Gated-Viewing Laser Imaging with Compressive Sensing},''
  \emph{Applied optics}, vol.~53, no.~33, pp. 7992--7997, 2014.

\bibitem{Andrea}
A.~Cola{\c{c}}o, A.~Kirmani, G.~A. Howland, J.~C. Howell, and V.~K. Goyal,
  ``{Compressive Depth map Acquisition using a Single Photon-Counting Detector:
  Parametric Signal Processing meets Sparsity},'' in \emph{Computer Vision and
  Pattern Recognition (CVPR), 2012 IEEE Conference on}.\hskip 1em plus 0.5em
  minus 0.4em\relax IEEE, 2012, pp. 96--102.

\bibitem{Sun}
M.-J. Sun, M.~P. Edgar, G.~M. Gibson, B.~Sun, N.~Radwell, R.~Lamb, and M.~J.
  Padgett, ``{Single-Pixel Three-Dimensional Imaging with Time-Based Depth
  Resolution},'' \emph{Nature communications}, vol.~7, p. 12010, 2016.

\bibitem{Epi1}
S.~Achar, J.~R. Bartels, W.~L. Whittaker, K.~N. Kutulakos, and S.~G.
  Narasimhan, ``{Epipolar Time-of-Flight Imaging},'' \emph{{ACM Transactions on
  Graphics (ToG)}}, vol.~36, no.~4, p.~37, 2017.

\bibitem{Epi2}
M.~O'Toole, S.~Achar, S.~G. Narasimhan, and K.~N. Kutulakos, ``{Homogeneous
  Codes for Energy-Efficient Illumination and Imaging},'' \emph{{ACM
  Transactions on Graphics (ToG)}}, vol.~34, no.~4, p.~35, 2015.

\bibitem{Epi3}
M.~O'Toole, J.~Mather, and K.~N. Kutulakos, ``{3D Shape and Indirect Appearance
  by Structured Light Transport},'' in \emph{Computer Vision and Pattern
  Recognition (CVPR), 2014 IEEE Conference on}.\hskip 1em plus 0.5em minus
  0.4em\relax IEEE, 2014, pp. 3246--3253.

\bibitem{Anscombe}
F.~J. Anscombe, ``{The Transformation of Poisson, Binomial and
  Negative-Binomial Data},'' \emph{Biometrika}, vol.~35, no. 3/4, pp. 246--254,
  1948.

\bibitem{Foi}
M.~M{\"{a}}kitalo and A.~Foi, ``{Optimal Inversion of the Anscombe
  Transformation in Low-Count Poisson Image Denoising},'' \emph{IEEE
  transactions on Image Processing}, vol.~20, no.~1, pp. 99--109, 2011.

\bibitem{DeadTime}
M.~L. Larsen and A.~B. Kostinski, ``{Simple Dead-Time Corrections for Discrete
  Time Series of Non-Poisson Data},'' \emph{Measurement Science and
  Technology}, vol.~20, no.~9, p. 095101, 2009.

\bibitem{Harmany}
Z.~T. Harmany, R.~F. Marcia, and R.~M. Willett, ``{This is SPIRAL-TAP: Sparse
  Poisson Intensity Reconstruction Algorithms-Theory and Practice},''
  \emph{IEEE Transactions on Image Processing}, vol.~21, no.~3, pp. 1084--1096,
  2012 (Note: the MATLAB codes of Ref.~\cite{System} and~\cite{First_Photon}
  are based on the code of this journal. We therefore include this journal in
  our reference list).

\end{thebibliography}

\end{document}